\documentclass{article}

\usepackage{arxiv}

\usepackage[utf8]{inputenc} 
\usepackage[T1]{fontenc}    
\usepackage{hyperref}       
\usepackage{url}            
\usepackage{booktabs}       
\usepackage{amsfonts}       
\usepackage{nicefrac}       
\usepackage{microtype}      
\UseRawInputEncoding
\usepackage{amsmath,amssymb}
\usepackage{subcaption}
\usepackage{graphicx}
\usepackage{float}
\usepackage{todonotes}
\usepackage{cite}
\usepackage{makecell}
\usepackage{tabularx}
\usepackage{multicol}

\usepackage[T1]{fontenc}
\usepackage{amsmath,amssymb,amsfonts}
\usepackage{mathrsfs}
\usepackage{marvosym}
\usepackage{caption}
\usepackage{subcaption} 
\usepackage{algorithm} 
\usepackage{algpseudocode} 
\usepackage{mathtools}
\usepackage{euscript}

\usepackage{blindtext}
\usepackage{float}

\usepackage{cite}  

\DeclarePairedDelimiter\abs{\lvert}{\rvert}%
\DeclarePairedDelimiter\norm{\lVert}{\rVert}%

\DeclareMathOperator{\rel}{rel}
\DeclareMathOperator{\AP}{AP}
\DeclareMathOperator{\CC}{CC}
\DeclareMathOperator{\D}{D}
\DeclareMathOperator{\E}{E}
\DeclareMathOperator{\HD}{HD}
\DeclareMathOperator{\2D}{2D}
\DeclareMathOperator{\mAP}{mAP}
\DeclareMathOperator{\NPQ}{NPQ}
\DeclareMathOperator{\DPQ}{DPQ}
\DeclareMathOperator{\NP}{NP}
\DeclareMathOperator{\opt}{opt}
\DeclareMathOperator{\PP}{P}
\DeclareMathOperator{\rand}{rand}
\DeclareMathOperator{\score}{User\  Score}

\newcommand{\overbar}[1]{\mkern 1.5mu\overline{\mkern-1.5mu#1\mkern-1.5mu}\mkern 1.5mu}

\renewcommand{\algorithmiccomment}[1]{//#1}
\newcommand{\hexagon}{\text{\HexaSteel}}

\DeclareRobustCommand\circledot{\mathbin{\ooalign{$\circledcirc$\cr\hidewidth$\bullet$\hidewidth}}}
\newcommand\smalldot{\begin{picture}(1,1)
\put(.5,.35){\circle*{.35}}
\end{picture}}
\newcommand\mediumdot{\begin{picture}(1,1)
\put(.5,.35){\circle*{.48}}
\end{picture}}
\newcommand\largedot{\begin{picture}(1,1)
\put(.5,.35){\circle*{.7}}
\end{picture}}

\makeatletter
\let\oldabs\abs
\def\abs{\@ifstar{\oldabs}{\oldabs*}}

\newcommand{\orcid}[1]{\href{https://orcid.org/#1}{\includegraphics[height=\fontcharht\font`\B]{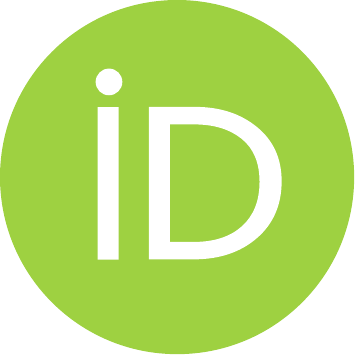}}}

\begin{document}

\title{Improved Evaluation and Generation of Grid Layouts using Distance Preservation Quality and Linear Assignment Sorting}

\author
{\parbox{\textwidth}{\centering 
K.\,U. Barthel \orcid{0000-0001-6309-572X},
        N. Hezel \orcid{0000-0002-3957-4672},
        K. Jung \orcid{0000-0002-3600-6848} and
        K. Schall \orcid{0000-0003-3548-0537}
        }
        \\
{\parbox{\textwidth}{\centering 
HTW Berlin, Visual Computing Group, Germany 
       }
}
}

\maketitle              
\begin{abstract}
Images sorted by similarity enables more images to be viewed simultaneously, and can be very useful for stock photo agencies or e-commerce applications. Visually sorted grid layouts attempt to arrange images so that their proximity on the grid corresponds as closely as possible to their similarity. Various metrics exist for evaluating such arrangements, but there is low experimental evidence on correlation between human perceived quality and metric value. We propose Distance Preservation Quality (DPQ) as a new metric to evaluate the quality of an arrangement. Extensive user testing revealed stronger correlation of DPQ with user-perceived quality and performance in image retrieval tasks compared to other metrics.
In addition, we introduce Fast Linear Assignment Sorting (FLAS) as a new algorithm for creating visually sorted grid layouts. FLAS achieves very good sorting qualities while improving run time and computational resources.
\keywords{Grid-based Image Sorting \and Visualization of Retrieval Results \and Linear Assignment Sorting \and  
Sorting and Searching \and 
Empirical Studies in Visualization \and 
Perceived Sorting Quality
}
\end{abstract}

\section{Introduction}

\begin{figure*}[htb]
  \centering
  \includegraphics[width=.3\linewidth]{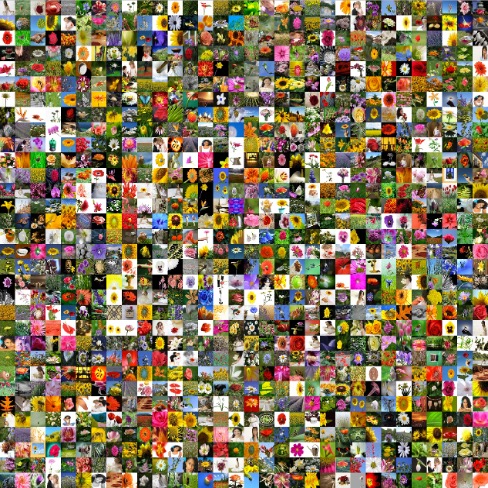}
  \hfill
  \includegraphics[width=.3\linewidth]{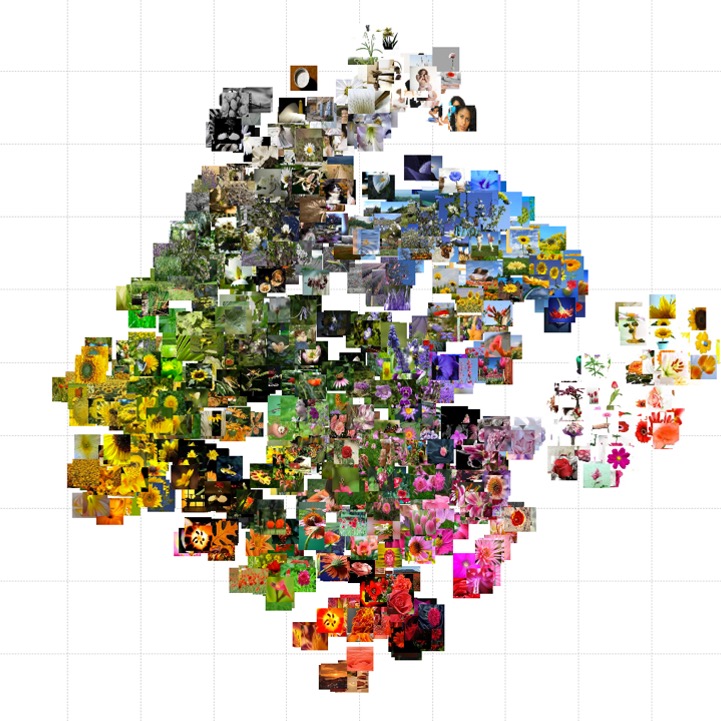}
  \hfill
  \includegraphics[width=.3\linewidth]{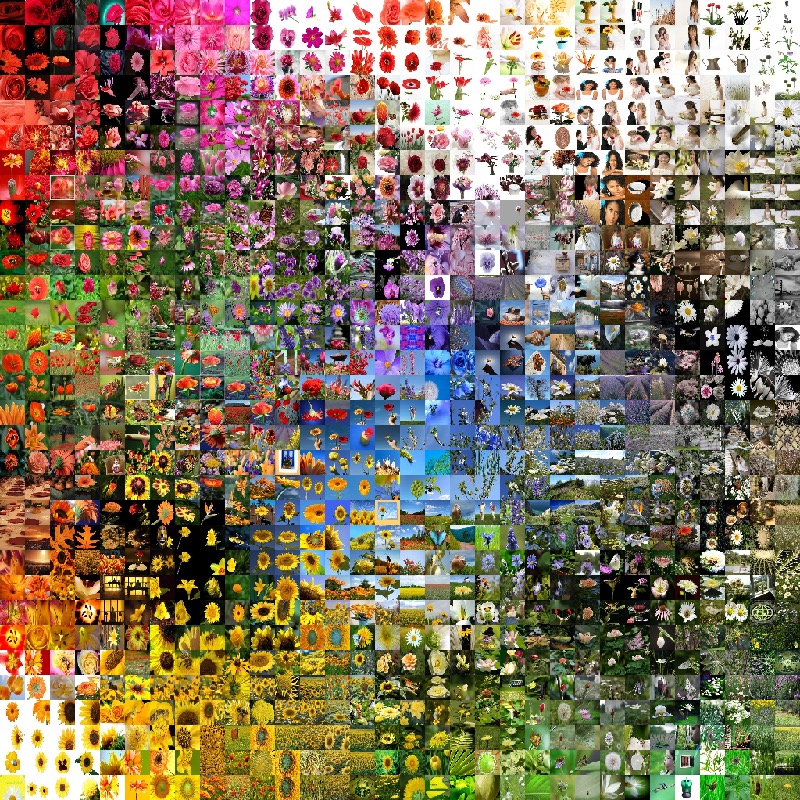}
  \caption{\label{fig:ex1}
           1024 "flower" images. Left: grid in random order. Center: t-SNE projection.
           Right: arranged with LAS.
}
\end{figure*}

It is difficult for humans to view large sets of images simultaneously while maintaining a cognitive overview of its content. 
As set sizes increase, the viewer quickly loses their perception of specific content contained in the set (Figure~\ref{fig:ex1} left). 
For this reason, most applications and websites typically display no more than 20 images at a time, which in many cases is only a tiny fraction of the images available. 
However, if the images are sorted according to their similarity, up to several hundred can be perceived simultaneously. It has been shown that a sorted arrangement helps users to identify regions of interest more easily and thus find the images they are looking for more quickly \cite{Schoeffmann2011, Quadrianto2010, journals/tog/ReinertRS13, journals/corr/HanZLXSM15}. The simultaneous display of larger image sets is particularly interesting for e-commerce applications and stock photo agencies.

In order to be able to sort images according to their similarity, a suitable measure of this similarity must be specified. 
Image analysis methods can generate visual feature vectors and image similarity is then expressed by the similarity of their feature vectors. 
While low-level feature vectors generated by classical image analysis techniques represent the general visual appearance of images (such as colors, shapes, and textures), vectors generated with deep neural networks can also describe the content of images \cite{Babenko2014, DeepFeatures, Radenovic2019, Cao2020}. 
The dimensions of these vectors are on the order of a few tens for low-level features, while deep learning vectors have up to thousands of dimensions.  

If the images are represented as high-dimensional (HD) vectors, their similarities can be expressed by appropriate visualization techniques. A variety of dimensionality reduction techniques have been proposed to visualize high-dimensional data relationships in two dimensions.
Often a distinction is made between methods that use vectors or pairwise distances. 
However, these methods can be converted from one another; pairwise distances can be calculated from the vectors, and the rows of a distance matrix can be used as vectors. 
Numerous techniques (\emph{Principal Component Analysis} (PCA) \cite{Pearson1901}, \emph{Multidimensional Scaling} (MDS) \cite{Sammon1969}, \emph{Locally Linear Embedding} (LLE) \cite{Roweis2000}, \emph{Isomap} \cite{Tenenbaum2000}, and others) are described in in \cite{Sarveniazi2014}. Other methods that work very well are \emph{t-Distributed Stochastic Neighborhood Embedding} \mbox{(t-SNE)} \cite{Maaten2008}, \emph{Uniform Manifold Approximation and Projection} (UMAP) \cite{McInnes2018} and \emph{Subset Embedding Networks} \cite{Xie2021}. 

Visualization is achieved by projecting the high-dimensional data onto a two-dimensional plane. However, all of the above techniques are of limited use if the images themselves are to be displayed. The center of Figure~\ref{fig:ex1} shows a \hbox{t-SNE} projection of the relative similarity of 1024 flower images. 
Due to the dense positioning of the projected images, some overlap and are partially obscured. Furthermore, only a fraction of the display area is used. 
\hbox{Using} techniques such as \emph{DGrid} \cite{Hilasaca2021} would solve the overlap problem, but would still not make best use of the available space.

To arrange or sort a set of images by similarity while maximizing the display area used, three requirements must be satisfied:
\hbox{1. The} images should not overlap. 2. The arrangement of the images should cover the entire display area. 3. The HD similarity relationships of the image feature vectors should be preserved by the 2D image positions.
Requirements 1 and 2 can only be met if the images are positioned on a rectangular grid. For the 3\textsuperscript{rd} requirement, the images have to be positioned such that their spatial distance corresponds as closely as possible to the high-dimensional distance of their feature vectors, despite the given grid structure. 

The \emph{Self-Organizing Map} is one of the oldest methods for organizing HD vectors on a grid \cite{Kohonen1982, Kohonen2013}. \emph{Self-Sorting Maps} \cite{Strong2011,Strong2014} are a more recent technique that orders images using a hierarchical swapping method. Other approaches first project the HD vectors to two dimensions, which are then mapped to the grid positions.
Various metrics exist for assessing the quality of such arrangements, but there is little experimental evidence of correlation between human-perceived quality and these metrics. 

In our paper we first describe other existing quality metrics for evaluating sorted grid layouts, then we give an overview of existing algorithms for generating sorted 2D grid layouts.
The key contributions of our work are: 1. Inspired by the \emph{\hbox{k-neighborhood} preservation index} \cite{Fadel2015}, we propose \emph{Distance Preservation Quality} as a new metric for evaluating grid-based layouts. 2. We then propose \emph{\hbox{Linear Assignment} Sorting}, an algorithm that very efficiently produces high quality 2D grid layouts. 3. We conducted an extensive user study examining different metrics and show that distance preservation quality better reflects the quality perceived by humans. 
We furthermore performed qualitative and quantitative comparisons with other sorting algorithms. 
In the last section, we show how to generate arrangements with special layout constraints with our proposed sorting method. 
This paper is based on our previous work \cite{doi:https://doi.org/10.1002/9781119376996.ch11}.

\section{Related Work}

\subsection{Quality Evaluation of Distance Preserving Grid Layouts}
\label{Quality Evaluation}

A high quality image arrangement is one that provides a good overview, places similar images close to each other, and images \hbox{being} searched can be found quickly. An evaluation metric expresses the quality of a sorted arrangement with a single number. This value should highly correlate with the quality perceived by humans. We review commonly used evaluation metrics and examine their properties and problems.

Grid-based arrangement of high-dimensional data $X$ consists of finding a mapping (a sorting function) $S: X \mapsto Y$ or $S: x_i \mapsto y_i$, where $x_i$ is the $i^{\:th}$  high-dimensional vector, whereas $y_i$ is the $i^{\:th}$ position vector on the grid in $\mathbb{R}^2$.
The distance between high-dimensional vectors is denoted by $\delta(\cdot, \cdot)$ whereas $\lambda(\cdot, \cdot)$ denotes the corresponding spatial distance of positions of the 2D grid. 

\begin{figure}[ht]
  \centering
  \includegraphics[width=.125\linewidth]{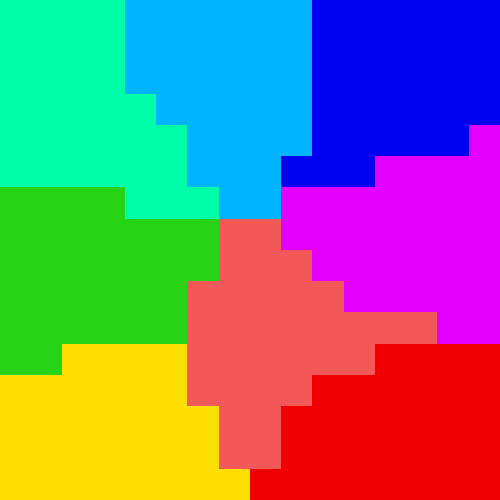} 
  \hspace{0.2\linewidth}
  \includegraphics[width=.125\linewidth]{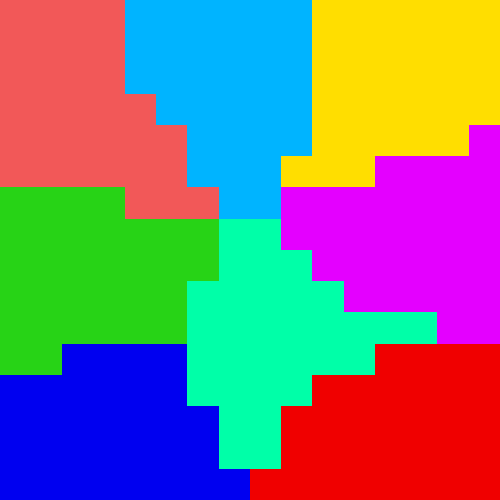}
  \caption{\label{fig:mAPExample}
           Two arrangements with the same mAP values, where people would rate the sorting quality unequally.}
\end{figure}

\textbf{Mean Average Precision.}
The Mean Average Precision (mAP) is the commonly used metric to evaluate image retrieval systems. 
\begin{equation} \label{eq:AP}
    \AP(q) = \frac{1}{m_q} \sum_{k=1}^{N} \PP_q(k) \rel_q(k)
\hspace{0.1\linewidth}
    \mAP = \frac{1}{N} \sum_{n=1}^N \AP(n)
\end{equation}
$\AP$ is the average precision, $N$ the number of total images, $m_q$ the number of positive images per class. $\PP_q(k)$ represents the precision at rank $k$ for the \hbox{query $q$}, $\rel_q(k)$ is a binary indicator function (1 if $q$ and the image at rank $k$ have the same class and 0 otherwise). 
The mAP metric defines a sorting as "good" if the nearest neighbors belong to the same class. In most cases, the mAP cannot be used because typically images do not have class information. Another problem is that mAP only considers images of the same class and ignores the order of the other images (see  Figure~\ref{fig:mAPExample}).

\textbf{k-Neighborhood Preservation Index.} 
The k-neighborhood preservation index $\NP_k(S)$ is similar to the mAP in that it evaluates the extent to which the neighborhood of the high-dimensional data set $X$ is preserved in the projected grid $Y$. It is defined as
\begin{equation} \label{eq:NPk}
\NP_k(S) = \frac{1}{N} \sum_{i=1}^{N} 
    \frac{\abs{\EuScript{N}_{k,i}^{\HD} \cap \EuScript{N}_{k,i}^{2\D}}} {k} 
\end{equation}
where $k$ is the number of considered neighbors, $\EuScript{N}_{k,i}^{\HD}$ is the set of the $k$ nearest neighbors of $x_i$ in the high-dimensional space, whereas $\EuScript{N}_{k,i}^{2\D}$ is the set of the $k$ nearest neighbors to $y_i$ on the 2D grid. 

The k-neighborhood preservation index has several problems: The quality of an arrangement is not described by a single value, but by individual values for each neighbor size $k$. Because of the discrete 2D grid, many spatial distances $\lambda$ are equal, which means there is no unique ranking of the grid elements. However, the biggest problem is a high sensitivity to noisy or similar distances.

\textbf{Cross-Correlation.} 
The cross-correlation is used to determine how well the distances of the projected grid positions correlate with the distances of the original vectors:
\begin{equation} \label{eq:CC}
\CC(S) = \sum_{i=1}^{N} \sum_{j=1}^{N} 
    \frac{(\lambda(y_i, y_j) - \Bar{\lambda}) 
          (\delta(x_i, x_j) - \Bar{\delta})} 
          {\sigma_\lambda \sigma_\delta} 
\end{equation}

The main problem of cross-correlation is that differences of large distances have a higher impact than differences of small distances. 
It may be problematic to assess the quality of an image arrangement with cross-correlation as it is equally important to maintain both small and large distances to keep similar images together and prevent dissimilar images from being arranged next to each other.

\textbf{Normalized Energy Function.} 
The normalized energy function measures how well the distances between the data instances are preserved by the corresponding spatial distances on the grid. 
\begin{equation} \label{eq:EF}
\begin{aligned}
\E_p(S) &= \min_{c}\biggl(\sum_{i=1}^{N} \sum_{j=1}^{N} 
    \frac{ \abs{c\cdot\delta(x_i, x_j) - \lambda(y_i, y_j)}^p} 
          { \sum_{r=1}^{N} \sum_{s=1}^{N} (\lambda(y_r, y_s))^p} \biggr)^\frac{1}{p}  
          \\
          \E_p'(S) &= 1 - \E_p(S)
\end{aligned}        
\end{equation}
The normalized energy function has essentially the same properties and problems as cross-correlation. The parameter $p$ can be used to tune the balance between small and large distances. Usually $p$ values of 1 or 2 are used
. Throughout this paper we use $\E_p'$ with a range of [0,1] with larger values representing better results. 

\subsection{Algorithms for Sorted Grid Layouts} \label{Algorithms_for_Sorted_Grid_Layouts}
\subsubsection*{Grid Arrangements} 

Since our new sorting method is based on both Self-Organizing Map and Self-Sorting Map, we present them here in more detail. 
 
A \textbf{Self-Organized Map} (SOM) uses unsupervised learning to produce a lower dimensional, discrete representation of the input space.
A SOM consists of a rectangular grid of map vectors $M$ having the same dimensionality as the input vectors $X$. 
To adapt a SOM for image sorting, the input vectors must all be assigned to different map positions, since multiple assignments would result in overlapping images.
Algorithm \ref{alg:SOM} describes the SOM sorting process.

\begin{algorithm}\small
	\caption{SOM}\label{alg:SOM} 
	\begin{algorithmic}[1]
	\State Initialize all map vectors with random values, 
	set learning rate $\alpha\  (<1) $ and neighbor radius
	
	\While {not convergence} \quad \Comment{ 
	convergence by reducing $\alpha$ and radius
	}
	    \ForAll {high-dimensional input vectors $x_i$}
	        \State Find the unassigned map position with most similar vector $m_j$ \par
            \State Assign the vector $x_i$ to this position and 
            \par
            \hskip\algorithmicindent update the neighbor map vectors: $m_{j'} = \alpha \cdot x_i + (1-\alpha)\cdot m_{j'}$
            
        \EndFor
        \State Reduce $\alpha$ and the neighbor radius
	\EndWhile
	\end{algorithmic} 
\end{algorithm}

A \textbf{Self-Sorting Map} (SSM) arranges images by initially filling cells (grid positions) with the input vectors. Then for sets of four cells, a hierarchical swapping procedure is used by selecting the best permutation from $4! = 24$ swap possibilities.
Algorithm \ref{alg:SSM} describes the sorting process with a SSM.
In \cite{Strong2013}, an alternative to SSMs is described that uses more sophisticated swapping strategies to achieve better global correlation, but at a much higher computational cost. 

\begin{algorithm}\small
	\caption{SSM}\label{alg:SSM}
	\begin{algorithmic}[1]
	\State Copy all input vectors into random but unique cells of the grid
	\State Divide the grid into 4x4 blocks
	\While {size of the blocks $\geq 1$}
	\State Divide each block into 2x2 smaller blocks 
	    \For {iteration $=1,2,\ldots L $ } \Comment{ $L$ = maximum number of iterations } 
	        \State For each block its target vector (the mean vector of its cells and adjacent blocks' cells) is calculated
	        \For {all blocks}
	        \For {all cells of the block}
        	    \State Find the best swapping permutation for the 4 cells from 
                corresponding positions of the adjacent 2x2 blocks\par
        	    \hskip\algorithmicindent \hskip\algorithmicindent \hskip\algorithmicindent by minimizing the sum of squared differences between
        	    the cell vectors and the target vectors of the blocks\par
                \EndFor
            \EndFor
        \EndFor
	\EndWhile
	\end{algorithmic} 
\end{algorithm}


\subsubsection*{Graph Matching} 
\textbf{Kernelized Sorting (KS)} \cite{Quadrianto2008} and \emph{Convex Kernelized Sorting} \cite{Djuric2012} generate distance-preserving lattices and find a locally optimal solution to a quadratic assignment problem \cite{Beckman1957}. KS creates a matrix of pairwise distances between HD data instances and a matrix of pairwise distances between grid positions. A permutation procedure on the second matrix modifies it to approximate the first matrix as well as possible, resulting in a one-to-one mapping between instances and grid cells.

\textbf{IsoMatch} also uses an assignment strategy to construct distance preserving grids \cite{Fried2015}. First, it projects the data into the 2D plane using the Isomap technique \cite{Tenenbaum2000} and creates a complete bipartite graph between the projection and the grid positions. 
Then, the Hungarian algorithm \cite{Kuhn1955} is used to find the optimal assignment for the projected 2D vectors to the grid positions. 
\hbox{IsoMatch} uses the normalized energy function $E_1$ trying to maximize the overall distance preservation. 

Similarly, DS++ presents a convex quadratic programming relaxation to solve this matching problem \cite{DS++}.
KS, IsoMatch and DS++ are not limited to rectangular grids. They can create layouts of any shape.
As with IsoMatch, any other dimensionality reduction methods (such as t-SNE or UMAP) can be used to first project the high-dimensional input vectors onto the 2D plane, and then re-arrange them on the 2D grid. 
A fast placing approach can be found in \cite{Hilasaca2021}. Any linear assignment scheme like the Jonker-Volgenant Algorithm \cite{Jonker1987} can be used to map the projected 2D positions to the best grid positions. 
Many nonlinear dimensionality reduction methods have been recently proposed, but the question of their assessment and comparison remains open. Methods comparing HD and 2D ranks are reviewed in \cite{Lee2008, Lueks2011}. 

\newcommand{\ts}{\textsuperscript}

\section{A New Quality Metric for Grid Layouts}
\label{sec:A New Quality Metric}
\begin{figure}[h]
\begin{center}
  \includegraphics[width=0.4\linewidth]{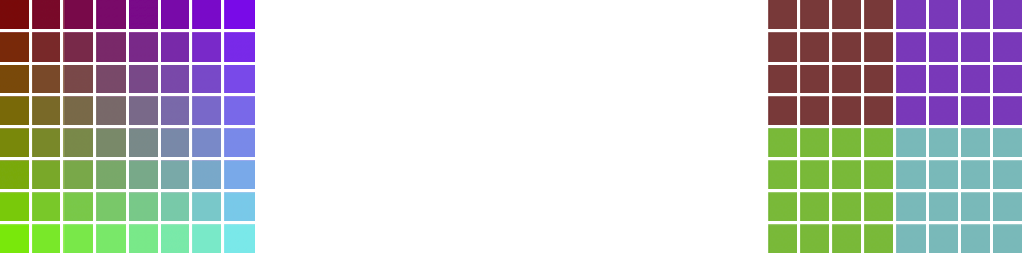}
\end{center}
  \caption{\label{fig:colors_sorting}
   Two examples of best possible 2D arrangements for 64 color samples. Left: The perfect HD (3D) order is preserved. Right: 4 different colors with 16 samples each, here the HD order cannot be preserved. Colors adjacent to a different color on the 2D grid are not near their nearest HD neighbors.}
\end{figure}

Our goal is to develop a metric that better reflects perceived quality. The quality is to be expressed with a single value, where 0 stands for a random and 1 for a perfect arrangement.  
There are two approaches when designing a suitable quality function for grid layouts. 
The first option would be to refer to the best possible 2D sorting that can theoretically be achieved. However, this approach is not applicable because the best possible sorting is usually not known.
The only viable way is to refer to the distribution of the high-dimensional data. A perfect sorting here means that all 2D grid distances are proportional to the HD distances. However, depending on the specific HD distribution, it is usually not possible to achieve this perfect order in a 2D arrangement (see Figure~\ref{fig:colors_sorting}).

\subsection{Neighborhood Preservation Quality} \label{Neighborhood Preservation}
Our initial approach towards a new evaluation metric was to combine the $k$-neighborhood preservation index values $\NP_k(S)$ to a single quality value. The $\NP_k(S)$ values for a perfect arrangement $S_{\opt}$ and the expected value for random arrangements $S_{\rand}$ are 
\begin{equation}
\NP_k(S_{\opt}) = 1 
\hspace{0.1\linewidth}
\mathbb{E}[\NP_k(S_{\rand})] =  \frac{k}{K} 
\end{equation}
where $k$ is the evaluated neighborhood size, $K$ is the maximum number of neighbors which is the number of HD data elements - 1. The expected $\NP_k$ value for a random arrangement is $\frac{k}{K}$, since more and more correct nearest neighbors are found as $k$ increases.

For a given 2D arrangement $S$ we define the \textit{Neighborhood Preservation Gain} $\Delta\NP_k^{\2D}(S)$ as the difference between the actual $\NP_k(S)$ value and the expected value for random arrangements.
\begin{equation}
\Delta\NP^{\2D}_k(S) = \max{(\NP_k(S)-\frac{k}{K},0)} 
\end{equation}
The maximum is taken because theoretically an arrangement can be worse than a random arrangement. This happens very rarely, but if it does, the negative values are very small. 
Since an optimal arrangement preserves all HD neighborhoods perfectly, we define
\begin{equation}
\Delta\NP^{\HD}_k = \Delta\NP^{\2D}_k(S_{\opt}) =  1-\frac{k}{K} 
\end{equation}

\begin{figure}[t]
\centering
  \includegraphics[width=0.6\linewidth]{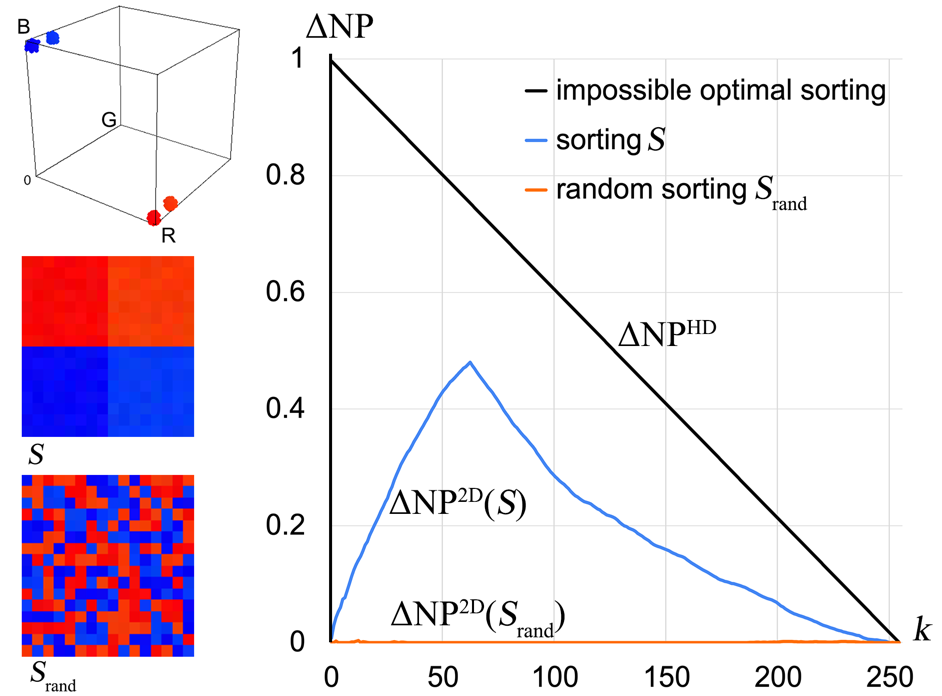}
  \caption{\label{fig:4colors_noise_deltaNP}
    Left: A 3D RGB histogram of a set of 64 x 4 colors, all slightly modified by noise. Below two arrangements of this set with sorted $(S)$ and random positions $(S_{\rand})$. Right: The corresponding curves of the neighborhood preservation gain $\Delta\NP$. Although the sorting $S$ is rather good, $\Delta\NP^{2D}(S)$ is very low for small $k$ values.}
\end{figure}

Figure~\ref{fig:4colors_noise_deltaNP} shows an example of 4 different primary colors, each used 64 times. All colors were slightly changed by some noise, resulting in 256 different colors. On the left side two arrangements and the color histogram are shown. The $\Delta\NP$ curves are shown on the right. Here the optimal HD order cannot be preserved in 2D.

We determine the vectors of neighborhood preservation gains of the actual 2D arrangement and of the perfect HD arrangement. We define the \emph{Neighborhood Preservation Quality} $\NPQ_p(S)$ as the \hbox{ratio} of the norms of these vectors.
$\NPQ_p(S)$ is close to 0 for a random and 1 for a perfect sorting. 
\begin{equation} \label{eq:NPQ}
\NPQ_p(S) = \frac{\norm{\Delta\NP^{\2D}(S)}{_p} }{\norm{\Delta\NP^{\HD}}{_p} } 
\hspace{0.1\linewidth}
 0 \leq \NPQ_p(S) \leq 1
\end{equation}

Each 2D grid position has several other positions with the same spatial distance $\lambda$ (e.g. the four nearest neighbors with a distance of 1). To determine $\NP_k(S)$ for all k-values, the mean k-neighborhood preservation index values must be determined for equal spatial distances. This ensures that isometries (rotated or mirrored arrangements) result in equal $\NPQ$ values.

To determine the neighborhood preservation quality $\NPQ_p(S)$, the value for the $p$-norm must be chosen. Higher values give more weight to $\NP_k$ values with smaller $k$, so the preservation of nearer neighbors becomes more important. In the case of very large $p$ values, only the four adjacent positions are taken into account. 

One problem with the proposed neighborhood preservation quality is its sensitivity to noisy distances in the HD data. 
This often occurs when using visual feature vectors. 
As image analysis is not perfect, a feature vector can be considered as a "perfect" vector that is disturbed by some noise. 
This effect can be seen in Figure~\ref{fig:4colors_noise_deltaNP}. \hbox{Although} sorting $S$ is rather good, the $\Delta\NP_k(S)$ values are very low, especially for near neighbors.
The top row of Figure~\ref{fig:NP_compare} shows the resulting order when ranking different arrangements of this data set according to their $\NPQ$ values for $p=2$. It can be seen that $\NPQ$ does not reflect the perceived sorting quality well.

\subsection{Distance Preservation Quality} 
The problem of the proposed neighborhood preservation quality consists of the fact that only the correct ranking of the neighbors is taken into account. The actual similarity of wrongly ranked neighbors is not considered. 
To address the noise-induced degradation of the neighborhood preservation quality, we propose not to compare the correspondence of the closest neighbors, but to compare the averaged distances of the corresponding neighborhoods $\EuScript{N}_{k,i}^{\HD}$ and $\EuScript{N}_{k,i}^{2\D}$.
For this, the \emph{average neighbor distances} for the $k$ closest neighbors are determined in HD and 2D: 
\begin{equation} \label{eq:D_HD}
\D_k^{\HD} = \frac{1}{kN} \sum_{i=1}^{N} \hspace{-0.05cm}\sum_{j\in {\EuScript{N}_{{k,i}}^{\HD}}}\hspace{-0.2cm} \delta(x_i, x_j)
\hspace{0.03\linewidth}
\D_k^{2\D}\hspace{-0.05cm}(S) = \frac{1}{kN} \hspace{-0.05cm}\sum_{i=1}^{N} \sum_{j\in {\EuScript{N}_{{k,i}}^{2\D}}} \hspace{-0.2cm}\delta(x_i, x_{j})
\end{equation}
It should be noted that the distances $\delta$ of the high-dimensional vectors are used for both the HD and the 2D neighborhoods. The only difference is that the sets of the actual $k$ nearest neighbors in HD and 2D are not the same if the 2D arrangement is not optimal.  

\begin{figure}[t]
  \includegraphics[width=0.6\linewidth]{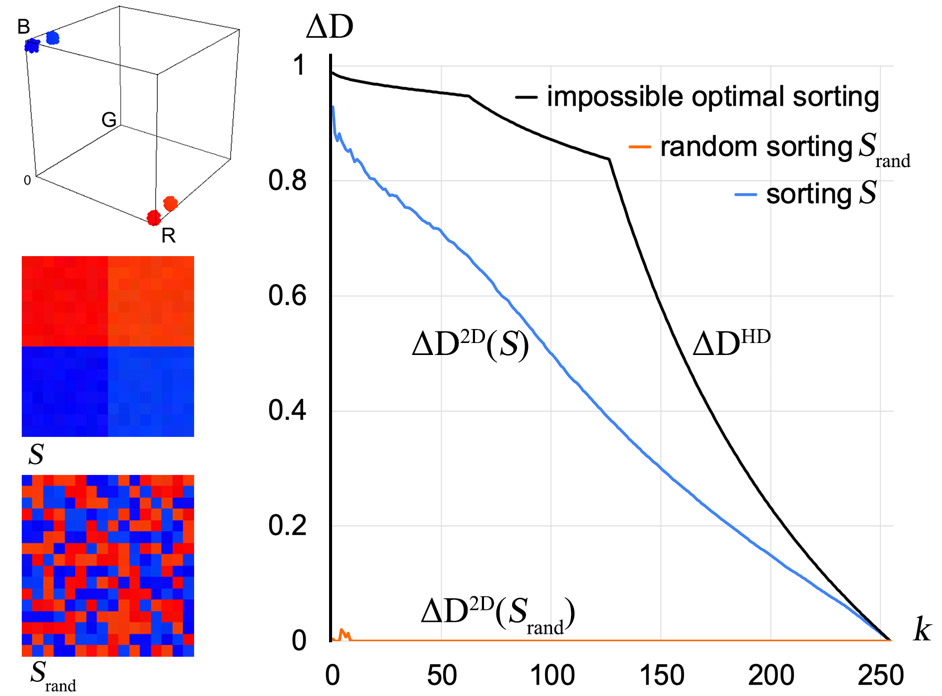}
  \centering
  \caption{\label{fig:4colors_noise_deltaDP}
    The same arrangements as the previous example. Here the curves show the distance preservation gain $\Delta\D$.}
\end{figure}

Similar to the neighborhood preservation quality, we compare the average neighborhood distance with the expectation value of the average neighborhood distance of random arrangements, which is equal to the global average distance $\overbar{\D}$ of all HD vectors $x_i$.
\begin{equation} \label{eq:D_mean}
\mathbb{E}[\D_k^{2\D}(S_{\rand})]\: =\: \overbar{\D} \:= \:\frac{1}{N^2} \sum_{i=1}^{N} \sum_{j=1}^{N} \delta(x_i, x_j)
\end{equation}
Analogous to $\Delta\NP_k^{\2D}(S)$, we define the \textit{Distance Preservation Gain} $\Delta\D_k$ as the difference between the average neighborhood distance of a random arrangement and the sorted arrangement.
\begin{equation} \label{eq:DP_k_HD}
\quad\Delta\D_k^{\HD} = \frac{1}{\overbar{\D}} (\overbar{\D} - \D_k^{\HD} ) 
\hspace{0.6cm}
\Delta\D_k^{\2D}(S) = \max{(\frac{1}{\overbar{\D}} (\overbar{\D} - \D_k^{\2D}(S)), 0)}    
\end{equation}
Compared to $\Delta\NP$, the order of subtraction is reversed for $\Delta\D$, since a higher distance is considered instead of a lower neighbor preservation.
Taking the difference between $\overbar{\D}$ and the average neighbor distance ensures $\Delta\D_k^{\2D}(S_{\rand})$ is approximately 0 for random arrangements. In theory, the division by $\overbar{\D}$ is not necessary, but limiting the values to a range from 0 to 1 improves the numerical stability when calculating the norm of the distance preservation gain for larger $p$ values.
Figure~\ref{fig:4colors_noise_deltaDP} shows the $\Delta\D$ curves of the previous example. 
It shows that for the sorted arrangement $S$, the $\Delta\D^{\2D}_k(S)$ values are much higher for small neighborhoods $k$, indicating that close neighbors on the grid are similar. Here the mean of HD distances for neighbors with equal 2D distances was used.  

The \emph{Distance Preservation Quality} $\DPQ_p(S)$ is defined as the ratio of the $p$-norms of the distance preservation gains of the actual arrangement to a perfect arrangement:
\begin{equation} \label{eq:DPQ}
\DPQ_p(S) = \frac{\norm{\Delta\D^{\2D}(S)}{_p} }{\norm{\Delta\D^{\HD}}{_p} } 
\hspace{0.1\linewidth}
 0 \leq \DPQ_p(S) \leq 1
\end{equation}
For a random arrangement, $\DPQ_p(S_{\rand})$ will be approximately 0, for a perfect arrangement $\DPQ_p(S_{\opt})$ will be 1. The influence of $p$ is evaluated in the user study section  \ref{User-Study}.

\begin{figure}[h!]
    \centering
  \includegraphics[width=0.6\linewidth]{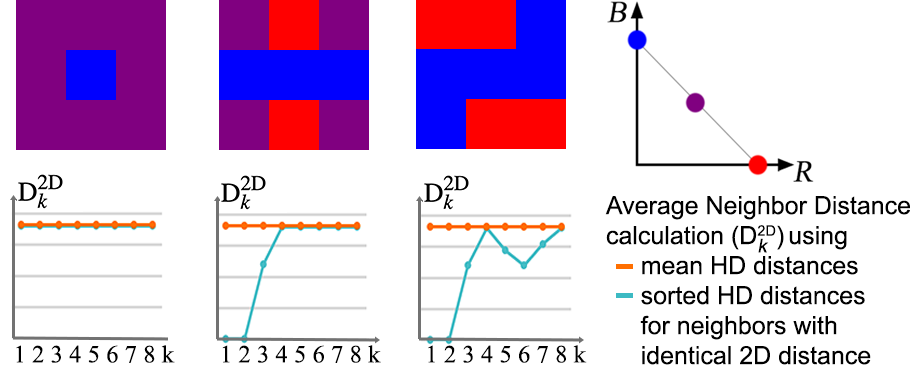}
  \caption{\label{fig:blau_rot_lila}
 Three neighbor constellations for the blue color in the center. The distance from blue to purple is equal to the mean of the distances to red and blue. Here $\D_k^{\2D}$  for $k = 1 ... 8$ is the same when using the mean HD distances for neighbors with equal 2D distances. Sorting these neighbors by their HD distance results in lower $\D_k^{\2D}$ values. This would better describe human perception if they preferred the right constellation to the left.}
\end{figure}

\begin{figure}[!h]
    \centering
  \includegraphics[width=.09\columnwidth]{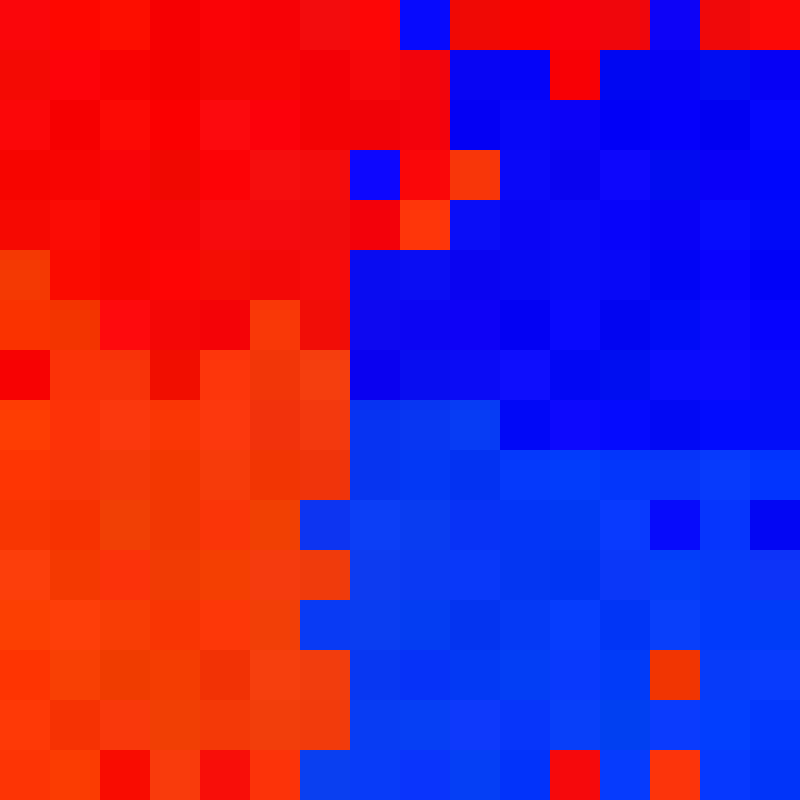} \hfill
  \includegraphics[width=.09\columnwidth]{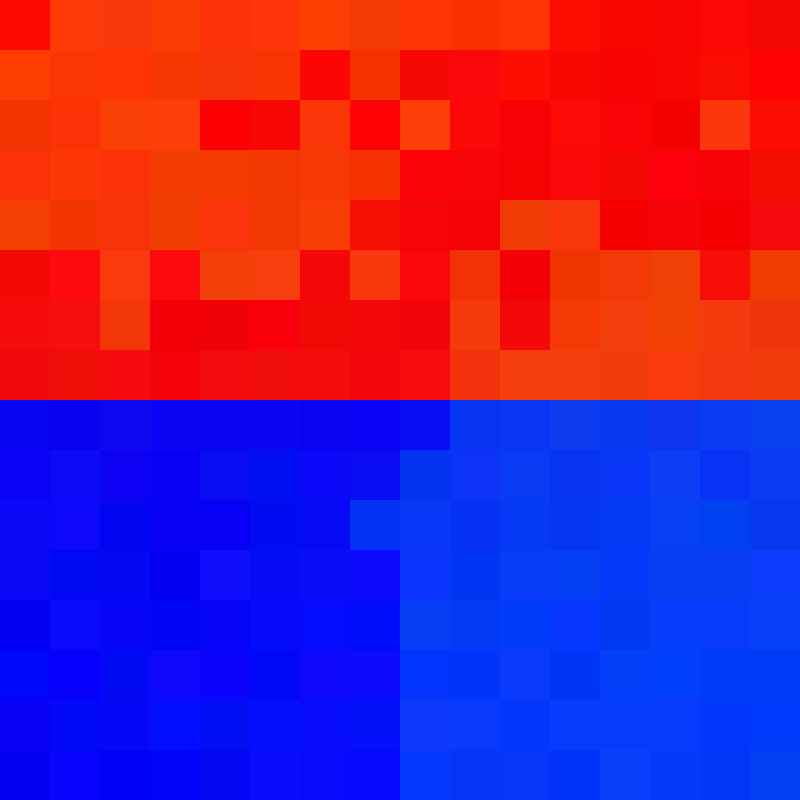} \hfill
  \includegraphics[width=.09\columnwidth]{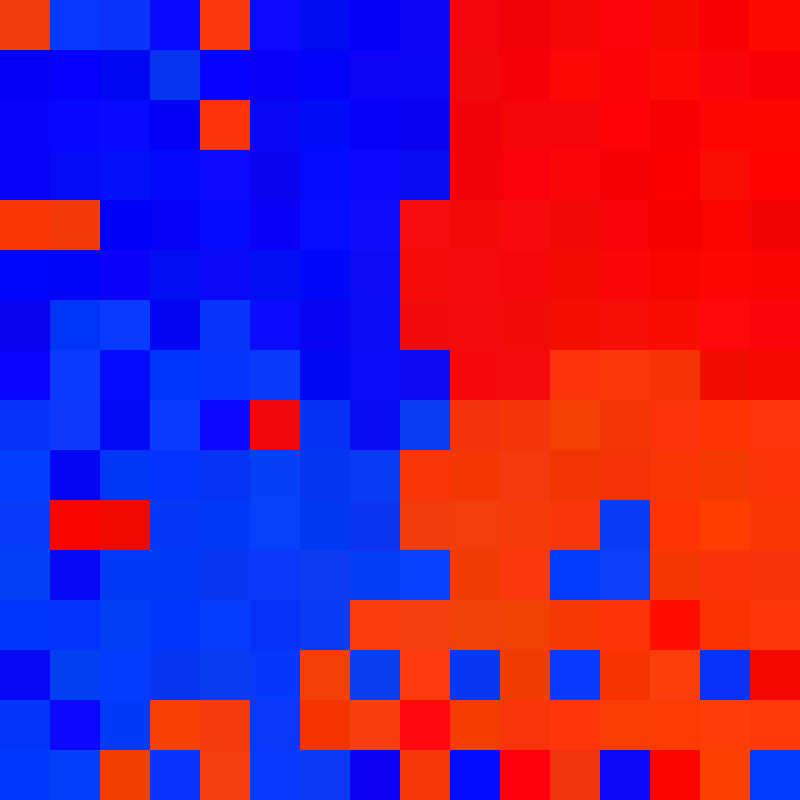} \hfill
  \includegraphics[width=.09\columnwidth]{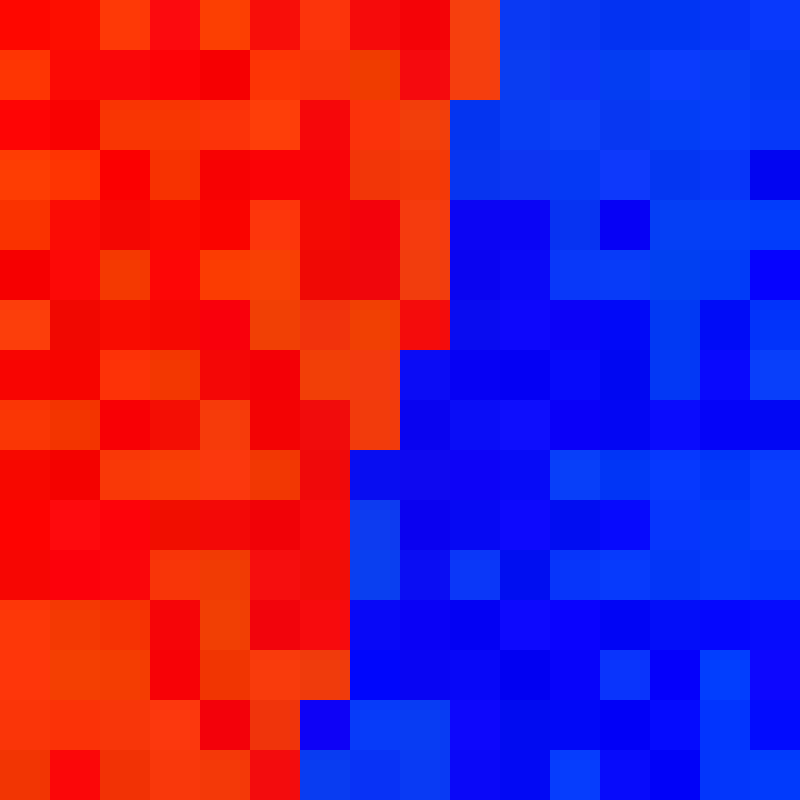} 
  \hfill
  \hfill
  \hfill
    \includegraphics[width=.09\columnwidth]{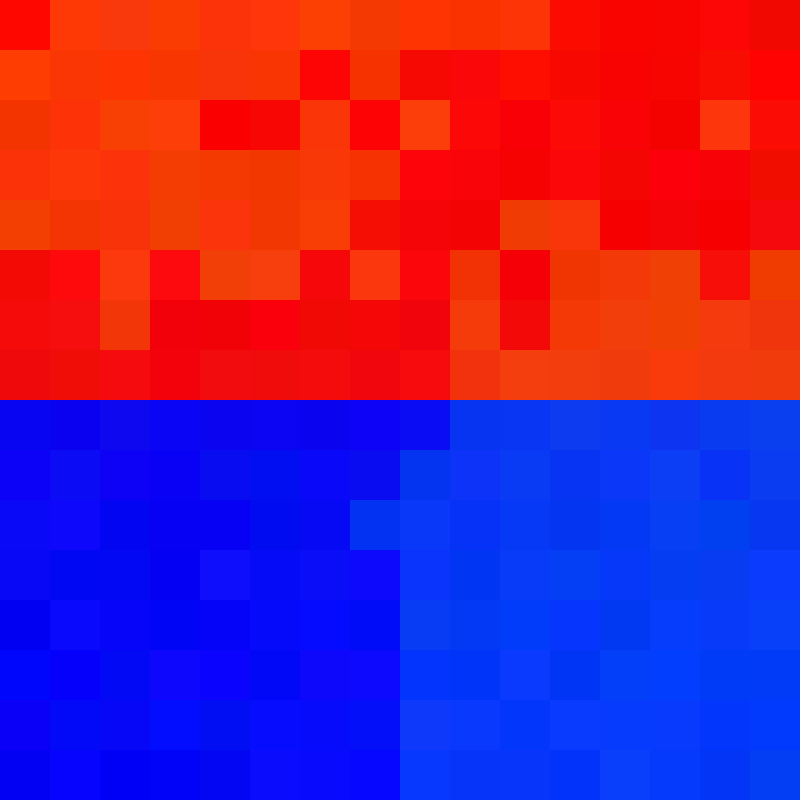} \hfill
  \includegraphics[width=.09\columnwidth]{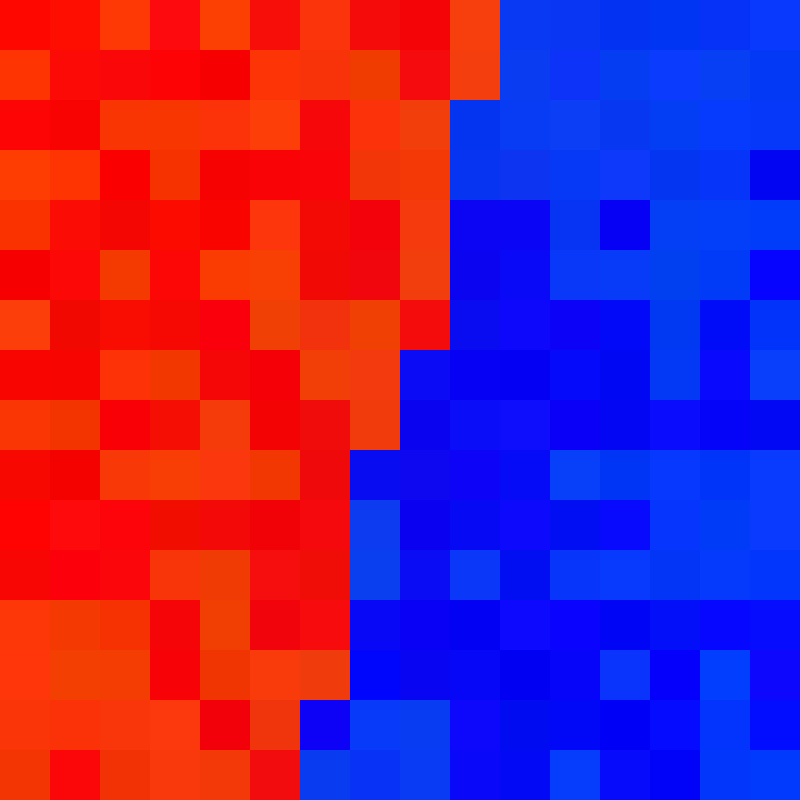} \hfill
  \includegraphics[width=.09\columnwidth]{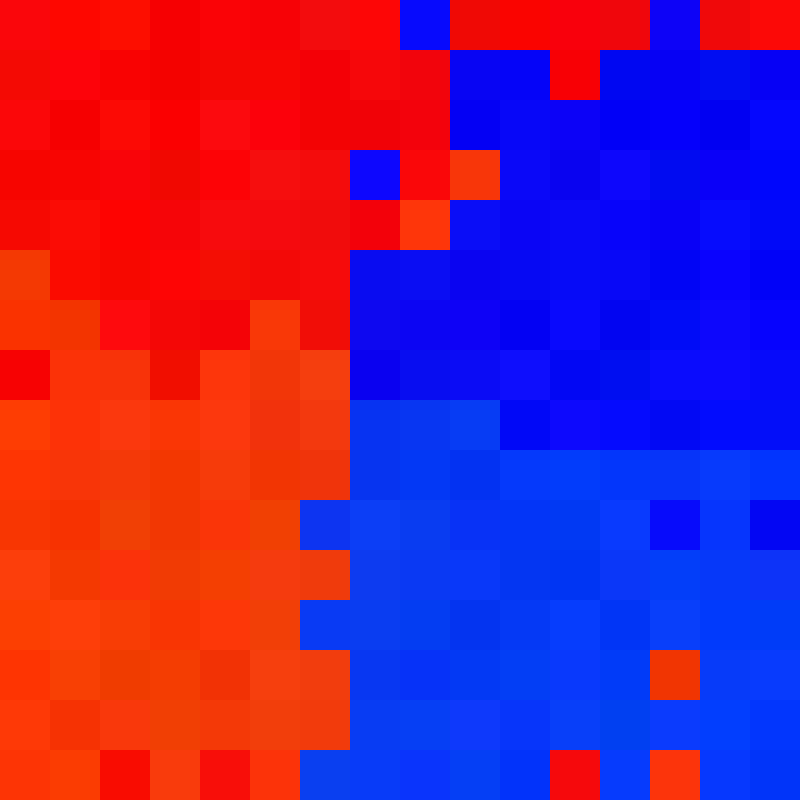} \hfill
  \includegraphics[width=.09\columnwidth]{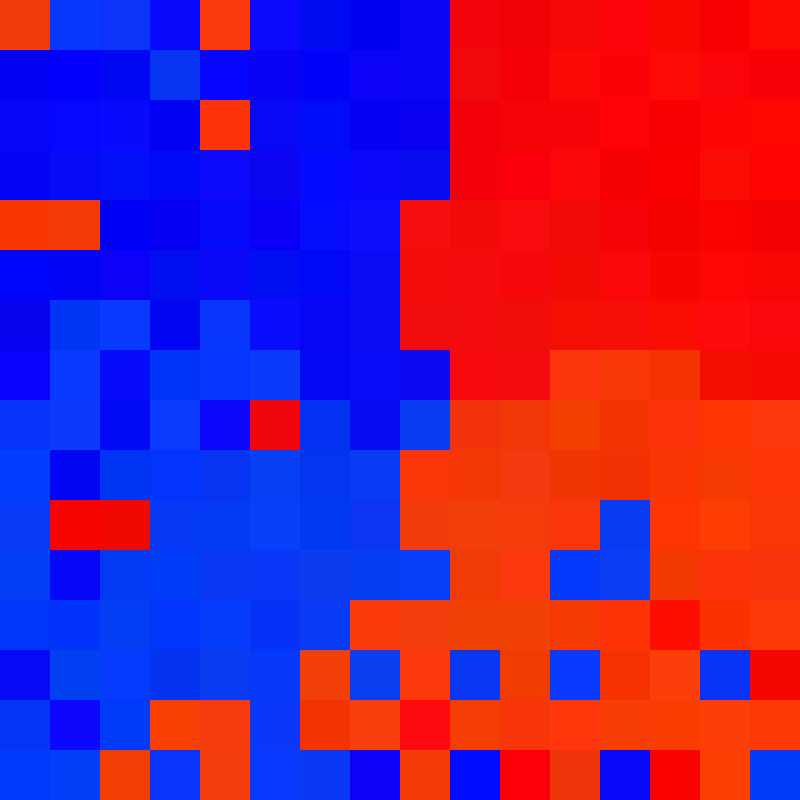} 
    \parbox[t]{\columnwidth}
    {\relax \small 
    NPQ: 0.406\quad>\quad0.400\qquad>\quad\quad 0.312\qquad >\qquad 0.294 \qquad\qquad\ \ \   
    DPQ: 0.689\quad>\quad0.644\qquad>\quad\quad 0.612\qquad >\qquad 0.479 }
  \caption{\label{fig:NP_compare}
    Arrangements in the order of $\NPQ_2$ and $\DPQ_2$.}
\end{figure}

Again, the problem of equal spatial distances must be considered when determining the average distances for the $k$ nearest neighbors on the grid. 
There are two ways to approach this: One is to use the mean HD distance for neighbors with equal 2D distance. 
The other possibility is to sort these neighbors by their HD distance. 
The former would be a pessimistic estimate of $\D_k^{\2D}$, whereas the latter would be an optimistic estimate (see Figure \ref{fig:blau_rot_lila}). 
The use of mean HD distances for equal 2D distances, is denoted as $DPQ_p^-$. Whereas $DPQ_p$ denotes the use of sorted HD distances.
Figure~\ref{fig:NP_compare} shows a better ranking of arrangements when evaluated with $\DPQ$ quality than with $\NPQ$ (for $p = 2$). 

\section{Our new Sorting Algorithm: Linear Assignment Sorting} 
First, we show how SOM and SSM can be optimized for speed and quality, which in combination leads to our new sorting scheme.

\subsection{Speed and Quality Optimizations of SOM and SSM}
The SOM described in section \ref{Algorithms_for_Sorted_Grid_Layouts} assigns each input vector to the best map vector and updates its neighborhood. 
The map update can be thought of as a blending of the map vectors with the spatially low-pass filtered assigned input vectors, where the filter radius corresponds to the current neighborhood radius. 
We propose to replace this time-consuming updating process: First, all input vectors are copied to the most similar unassigned map vector. Then, all map vectors are spatially filtered using a box filter.
It is possible to achieve constant complexity independent of the kernel size by using uniform or integral filters. \cite{Lewis1995, Viola2001}.
 Due to the sequential process of the SOM, the last input vectors can only be assigned to the few remaining unassigned map positions. This results in isolated, poorly positioned vectors.
 
The SSM avoids the problem of isolated, bad assignments by swapping the assigned positions of four input vectors at a time. 
To find the best swap, the SSM uses a brute force approach that compares the four input vectors with the four mean vectors of the blocks to which each swap candidate belongs. 
Due to the factorial number of permutations, adding more candidates would be computationally too complex. 
In order to still be able to use more swap candidates, we propose optimizing the search for the best permutation by linear programming. 
Another problem of the SSM is the use of a single mean vector per block, which incorrectly implies that all positions in the block are equivalent when they are swapped. 
The usage of a single mean vector per block can be considered as a subsampled version of the continuously filtered map vectors. 
Therefore, we propose using map filtering without subsampling, as this allows a better representation of the neighborhoods of the map vectors. 
The block sizes of the SSM remain the same for multiple iterations, this can be seen as repeated use of the same filter radius. 
We propose continuously reducing the filter radius.

\subsection{Linear Assignment Sorting} \label{Linear Assignment Sorting}
Our proposed new (image) sorting scheme called \textit{Linear Assignment Sorting} (LAS) combines ideas from the SOM (using a continuously filtered map) with the SSM (swapping of cells) and extends this to optimally swapping all vectors simultaneously.
The principle of the LAS algorithm can be described as follows: Initially all map vectors are randomly filled with the input vectors. Then, the map vectors are spatially low-pass filtered to obtain a smoothed version of the map representing the neighborhoods. In the next step all input vectors are assigned to their best matching map positions. This is done by finding the optimal solution by minimizing the cost $C$: 
\begin{equation}
\label{eq:optimal assignment}
\begin{gathered}
C = \sum^N_i \sum^N_j a_{ij}c_{ij} 
\hspace{0.05\linewidth}
\textrm{with}  
\hspace{0.05\linewidth}
a_{ij} \in \{0, 1\},
\hspace{0.05\linewidth}
c_{ij} = \norm{x_i-m_j}{^q}
\\
\textrm{subject to} 
\hspace{0.05\linewidth}
\sum^N_j a_{ij} = 1, 
\hspace{0.05\linewidth}
\sum^N_i a_{ij} = 1   
\end{gathered}
\end{equation}
$a_{ij}$ is a binary assignment value, whereas $c_{ij}$ is the distance between the input vectors $x_i$ and the map vectors  $m_j$. The power $q$ allows the distances to be transformed in order to balance the importance of large vs small distances. 
Since the number of possible mappings is factorial, we use the Jonker-Volgenant linear assignment solver \cite{Jonker1987} to find the best swaps with reduced complexity $\mathcal{O}(N^3)$. 
The actual sorting is achieved by repeatedly assigning the input vectors and filtering the map vectors with a successively reduced filter radius. 
The principle of the LAS sorting scheme for a grid of size \hbox{$N = W \cdot H$} is summarized in algorithm \ref{alg:LAS}: 
\begin{algorithm}[ht]\small
	\caption{LAS}\label{alg:LAS}
	\begin{algorithmic}[1]
	\State Set  \ \ $r_f = \lfloor max(W,H)\cdot f_{r0} \rfloor$\qquad\quad   //  initial filter radius ($f_{r0} \leq 0.5$)  
	\par $f_{r}$ \qquad\qquad\qquad\qquad\qquad\qquad\ \.  // radius reduction factor ($f_{r} < 1$)
	\State Assign and copy all input vectors to random but unique map vectors
	\While {$r_f > 1$}
	    \State Filter the map vectors using the actual filter radius $r_f$
	    \State Find the optimal assignment for all input vectors (acc. to Eq. \ref{eq:optimal assignment})
	    \State Copy all input vectors to the map vectors of their new positions
	    \State Reduce the filter radius: $r_f = r_f \cdot f_r$
	\EndWhile
	\end{algorithmic} 
\end{algorithm}

The only parameters of the LAS algorithm are the initial filter radius and the radius reduction factor, which controls the exponential decay of the filter radius and thus the quality and/or the speed of the sorting.
Examining different $q$ values for transforming the distances between the input and map vectors did not reveal much difference; in the interest of faster computations, we use $q=2$. 

Linear Assignment Sorting is a simple algorithm with very good sorting quality (see next section for results). 
However, for larger sets in the range of thousands of images, the computational complexity of the LAS algorithm becomes too high. 
However, with a slight modification of the LAS algorithm, very large image sets can still be sorted. 
\emph{Fast Linear Assignments Sorting} (FLAS) is able to handle larger quantities of images by replacing the global assignment with multiple local swaps, as described in Algorithm \ref{alg:FLAS}. This approach allows much faster sorting while having little impact on the quality of the arrangement. Comparisons between LAS and FLAS are given in the next section.

\begin{algorithm}[h]\small
	\caption{FLAS}\label{alg:FLAS} 
	\begin{algorithmic}[1]
	\State Set  \ \ $r_f = \lfloor max(W,H)\cdot f_{r0} \rfloor$  \quad//  initial filter radius ($f_{r0} \leq 0.5$)
	\par $f_{r}$ \qquad\qquad\qquad\qquad\qquad\ \   \algorithmiccomment{ radius reduction factor ($f_{r} < 1$) }
	\par $n_c$ \qquad\qquad\qquad\qquad\qquad\ \  \algorithmiccomment{ number of swap candidates}
	\par $iterations = W \cdot H \ /\  {n_c}$ 
	\par
	\State Assign and copy all input vectors to random but unique map vectors
	\While {$r_f > 1$}
	    \State Filter the map vectors using the actual filter radius
	    \For {$i = 1,2,\ldots iterations$} 
    	    \State Select a random position \& select $n_c$ random swap candidates
    	    \par \quad\; 
    	    (assigned input vectors) within a radius of $max(r_f, \frac{\sqrt{n_c} - 1}{2})$  
    	    \State Find the best swapping permutation
    	    \State 
    	    Assign the input vectors to their new map positions
        \EndFor
    \State Copy the input vectors to the map vectors of their assigned positions 
    \State Reduce the filter radius: $r_f = r_f \cdot f_r$
	\EndWhile
	\end{algorithmic} 
\end{algorithm}

The selection of FLAS parameters allows the control of the quality and speed of the sorting process. In this way, we generated many sorted arrangements of different quality, which were then used in the user study in Section \ref{User-Study}.

A sample implementation of the LAS and FLAS algorithms and the distance preservation quality $\DPQ_p$ can be found at\\ {\small \url{https://github.com/Visual-Computing/LAS_FLAS}}

\section{User Study} \label{User-Study}

\subsection{Experiment Design}
To evaluate the proposed $\DPQ_p$ metric and the new sorting schemes (LAS \& FLAS), an extensive user study was conducted. In a first experiment, we determined the correlation between user preferences and the  quality metrics described in Section \ref{Quality Evaluation} and \ref{sec:A New Quality Metric}. In a second experiment, we examined the relationship between the time required to find images in arrangements and the metrics' quality scores and the users' ratings, respectively.

\subsubsection*{Image Sets } 
\begin{figure*}[t!] 
    \captionsetup{singlelinecheck = false} 
    \begin{subfigure}{0.24\textwidth}
        {\includegraphics[width=\textwidth]{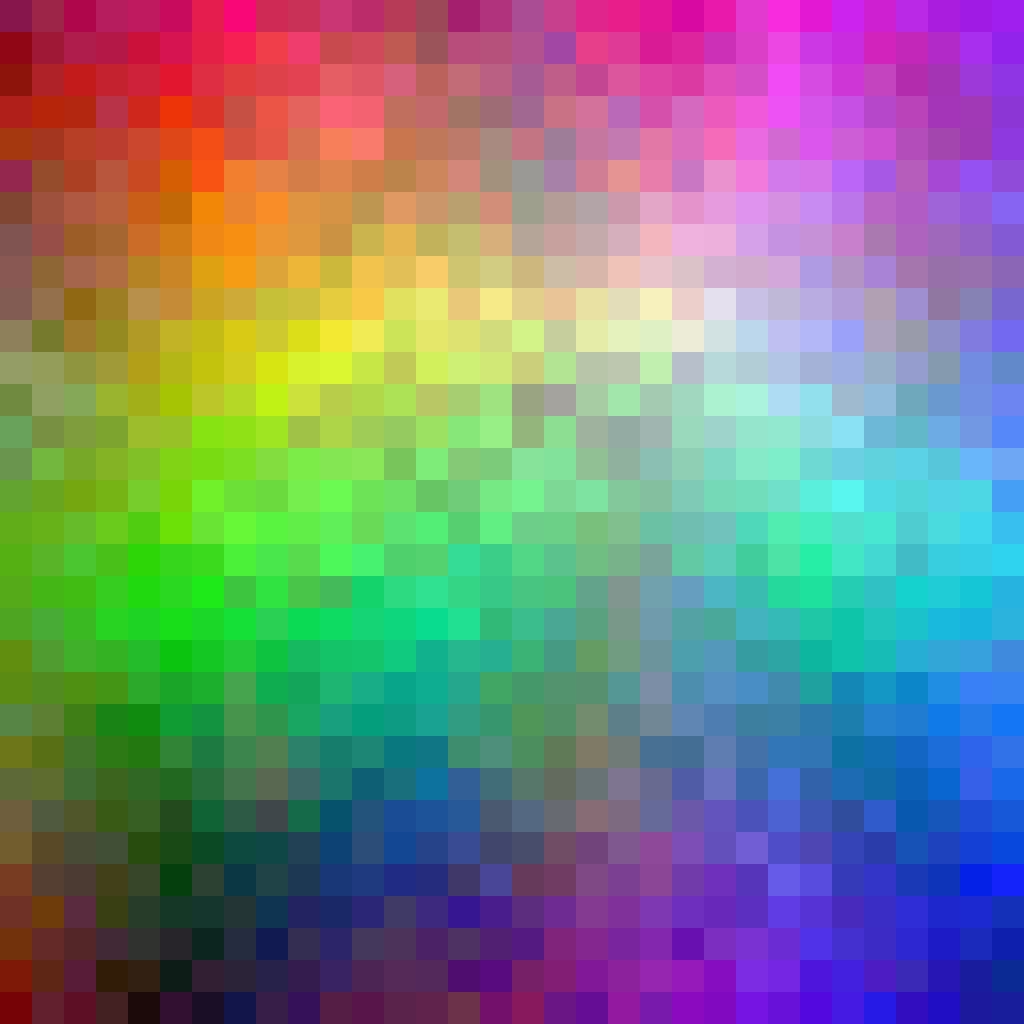}}
        \caption{1024 RGB colors}
    \end{subfigure}
    \hfill
    \begin{subfigure}{0.24\textwidth}
        {\includegraphics[width=\textwidth]{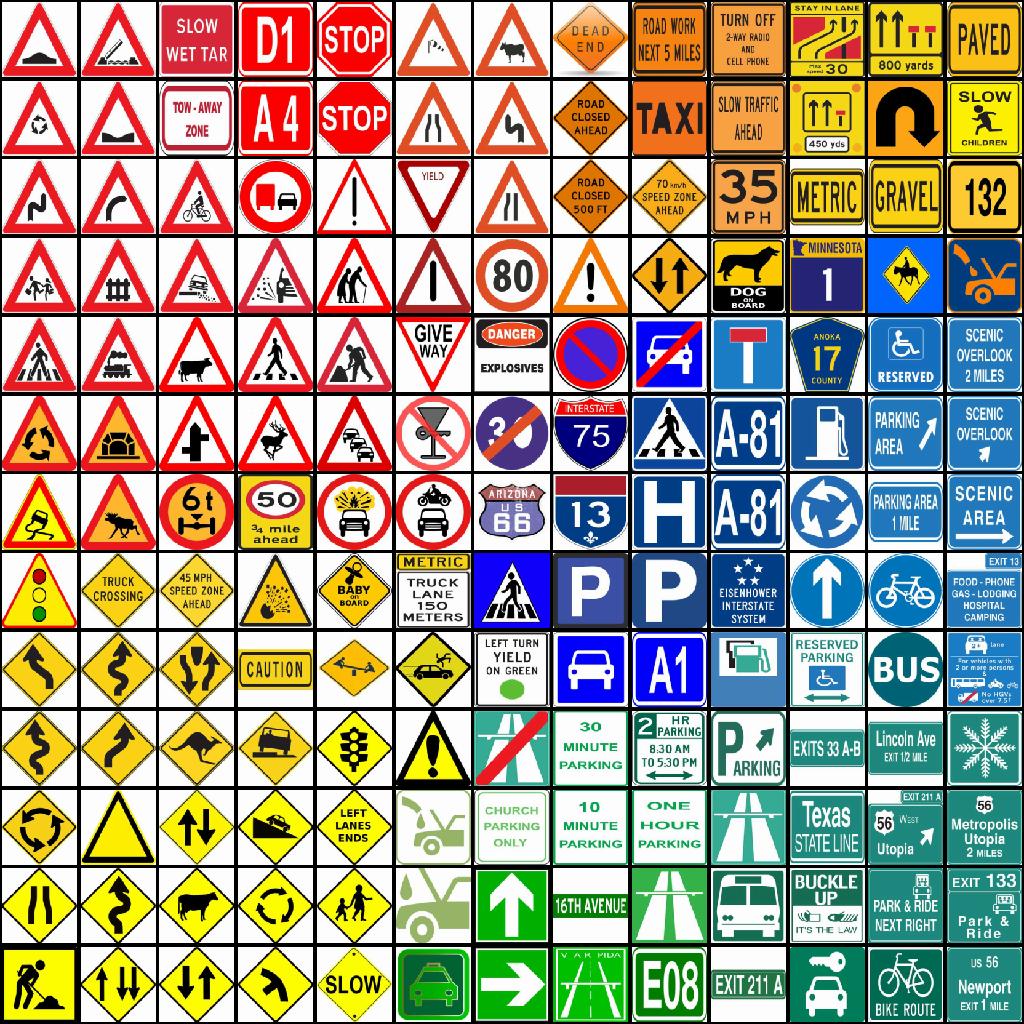}}
        \caption{ 169 traffic sign images}
    \end{subfigure}
    \hfill
    \begin{subfigure}{0.24\textwidth}
        {\includegraphics[width=\textwidth]{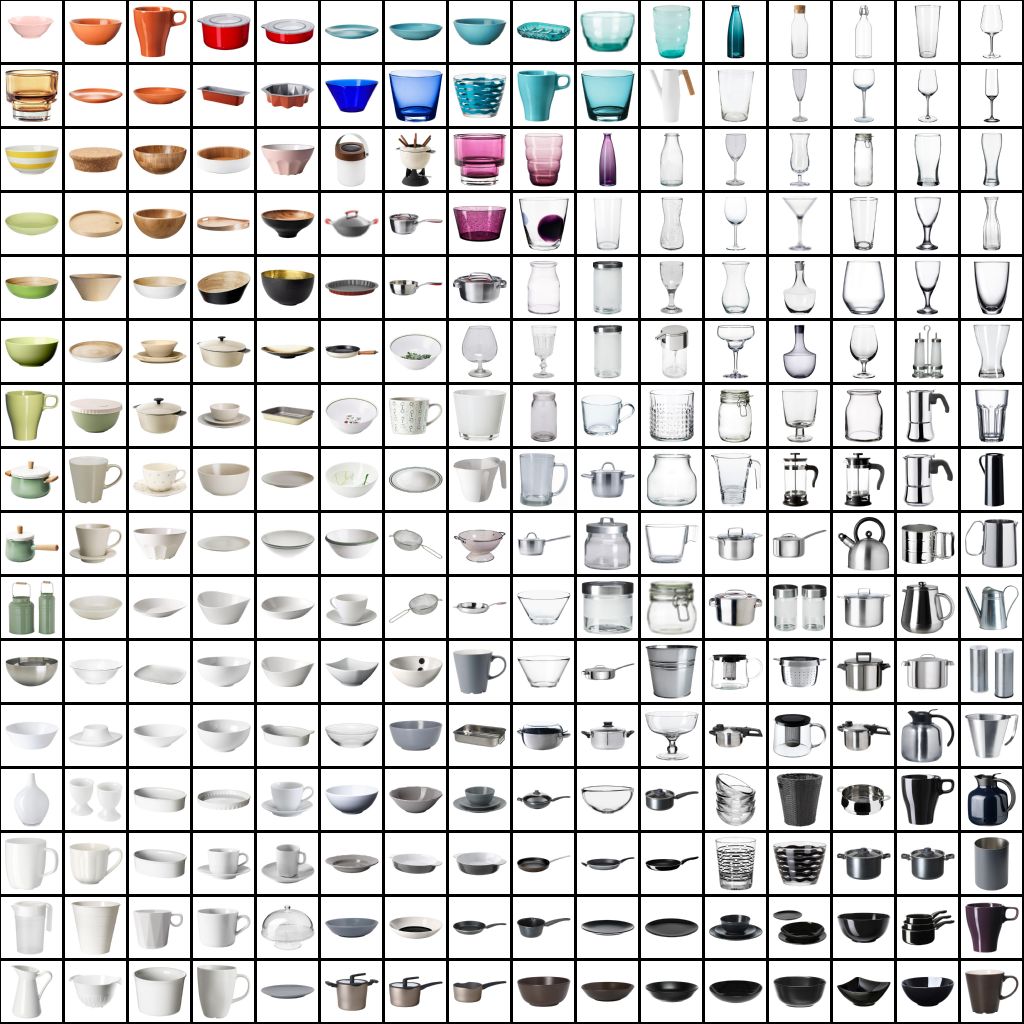}}
        \caption{ 256 kitchenware images }
    \end{subfigure}
    \hfill
    \begin{subfigure}{0.24\textwidth}
        {\includegraphics[width=\textwidth]{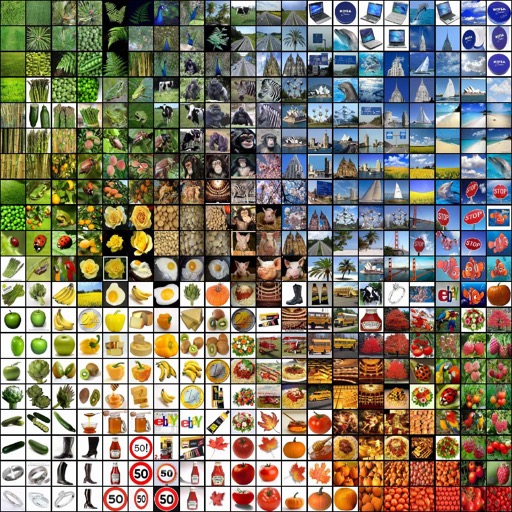}}
        \caption{ 400 images from the web}
    \end{subfigure}
    \caption{ \label{fig:image_testsets}
    The four sets of images used. The arrangements shown are the ones that received the highest user scores.}
\end{figure*}

 Figure \ref{fig:image_testsets} shows the four image sets used in the experiments. The first set consists of 1024 random RGB colors. The random selection implies that there is no specific low-dimensional embedding that can be exploited to project the data to 2D. While the RGB color set is a somewhat artificial set, we also used three sets of images that represent different scenarios. The first image set consists of 169 traffic sign images from the Pixabay stock agency. This is an example of images with a common theme that consist of multiple groups of visually similar images. The second set consists of 256 images of Ikea kitchenware items. These are the kind of images one might find on an e-commerce website. 
 Some of these images are very similar, which makes them difficult to be found in such a large collection. 
 The last set consists of 400 images of 70 unrelated concepts manually crawled from the web.

\subsubsection*{Feature Vectors}
For the RGB color set, the R, G, and B values were taken directly as vectors. Theoretically, the Lab color space would be better suited for human color perception, but even the Lab color space is not perceptually uniform for larger color differences. To ensure easy reproducibility of the results, we kept the RGB values. 

\begin{figure}[h]
    \centering
        {\includegraphics[width=0.24\textwidth]{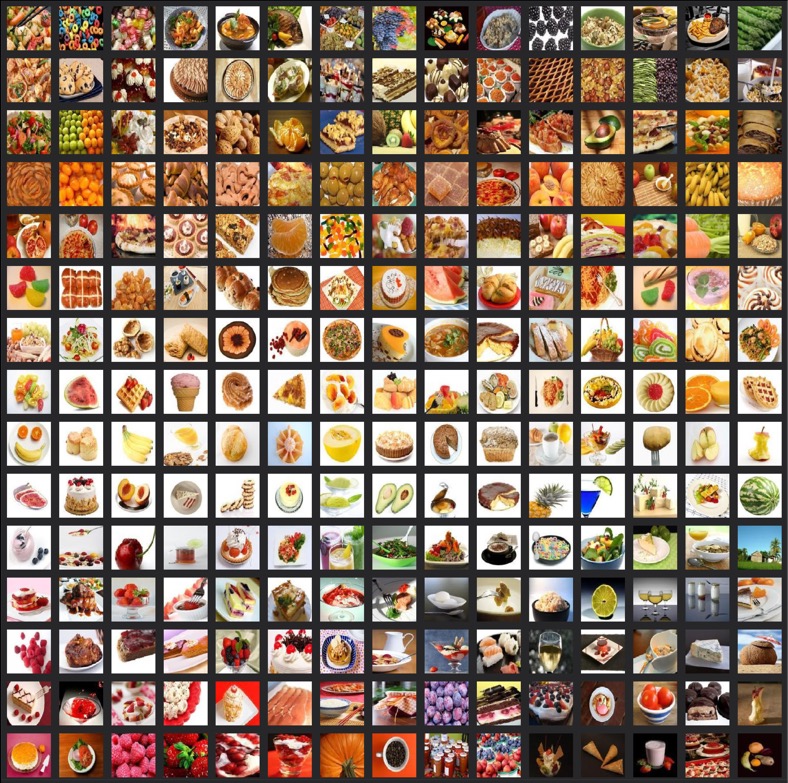}}
    \hspace{1cm}
        {\includegraphics[width=0.24\textwidth]{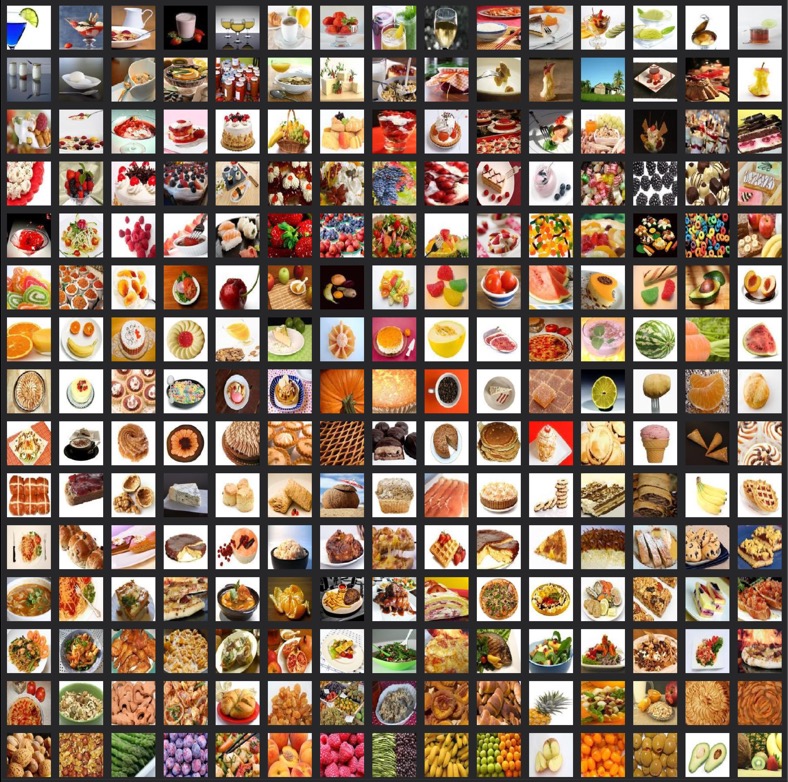}}
    \caption{\label{fig:low-level_vs_neural} Left: 256 images tagged with "food" sorted with low-level feature vectors. Right: the same images sorted with feature vectors from a neural network (MobileNetV3 small \cite{MobileNet}).
 } 
\end{figure}

For the three image sets, one might expect that high-dimensional feature vectors from deep neural networks would be best suited to describe these images, which is definitely true for retrieval tasks.
However, when neural feature vectors are used to visually sort larger sets of images, the arrangements often look somewhat confusing because images can have very different appearances even though they represent a similar concept (see Figure \ref{fig:low-level_vs_neural}).
Since people pay strong attention to colors and visually group similar-looking images when viewing larger sets of images, feature vectors describing visual appearance are usually more suitable for obtaining arrangements that are perceived as "well organized". For this reason, in our experiment, we used 50 dimensional low-level features similar to MPEG-7 features that describe the color layout, color histogram, and edge histogram of the images. 
However, the choice of feature vectors has limited impact on the experiments performed, since all sorting methods use the same feature vectors and the metrics indicate how well the similarities are preserved.

\subsubsection*{Implementation}
We organized the experiment as online user tests, where participants could take part in a raffle after completing the experiments. It was possible to perform the experiment more than once, but it was ensured that participants would not see the same arrangements twice. In total, more than 2000 people participated in the study.

\subsubsection*{Investigated Sorting Methods and Metrics}
In our experiments, we used sorted arrangements generated with the following methods: 
SOM, SSM, IsoMatch, LAS, FLAS, and the t-SNE 2D projection that was mapped to the best 2D grid positions (indicated as \hbox{t-SNEtoGrid)}. 
Several of these generated arrangements were then selected based on the range of variation in sorting results per method. 
The UMAP method was not investigated because in many cases its KNN graph broke into multiple components, which made an arrangement onto the 2D grid impossible.
In order to also have examples of low quality for comparison, some sorted arrangements were generated with FLAS using poor parameter settings (indicated as Low Qual.).

The evaluated quality metrics were the Energy function $E'_1$ and $E'_2$ (Equation \ref{eq:EF}) and the distance preservation quality $\DPQ_p(S)$ (Equation \ref{eq:DPQ}) with different $p$ values.
As the normalized energy function $\E_2'$ and cross-correlation provide an almost identical quality ranking for different arrangements we did not evaluate the cross-correlation metric.

\subsection{Evaluation of User Preferences}
In the first experiment, pairs of sorted image arrangements were shown. Users were asked to decide which of the two arrangements they preferred in the sense that "the images are arranged more clearly, provide a better overview and make it easier to find images they are looking for".

\begin{figure}[!htb]
    \centering
    \includegraphics[width=0.55\linewidth]{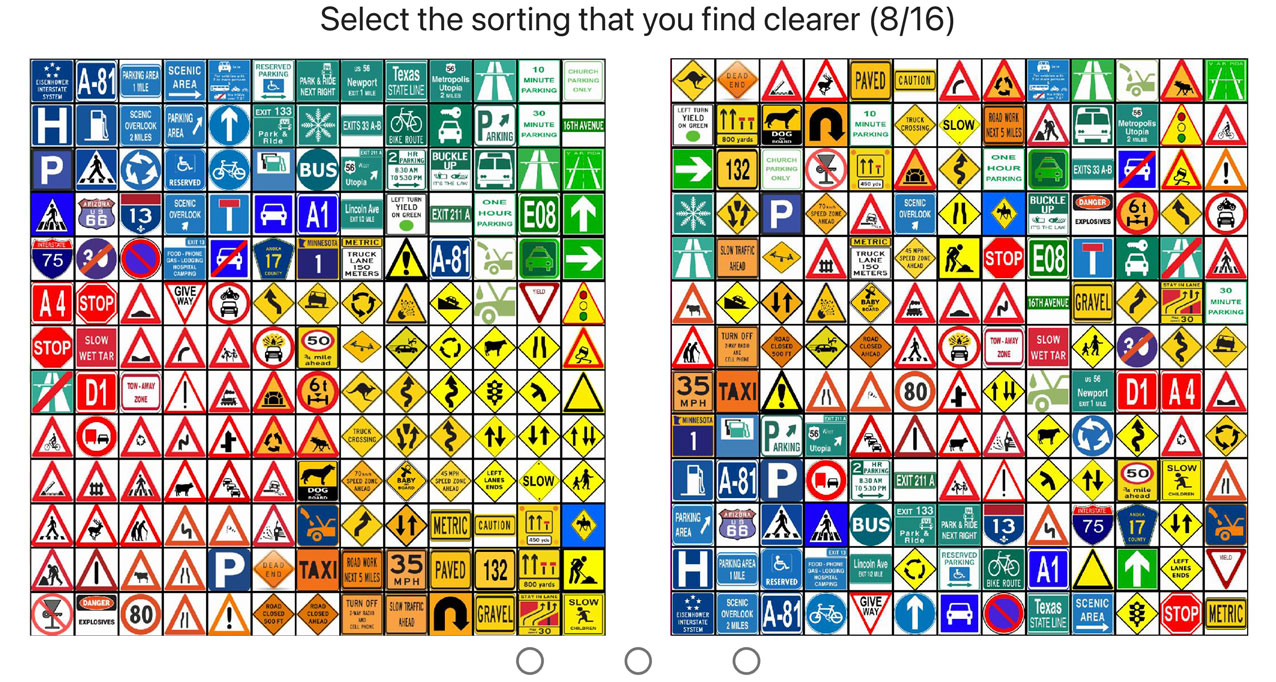}
    \caption{ User preferences. In this study we ask users to choose the arrangement they preferred.
    }
    \label{fig:compare_experiment}
\end{figure}

Figure \ref{fig:compare_experiment} shows a screenshot of this experiment. All users had to evaluate 16 pairs and decide whether they preferred the left or the right arrangement. They could also state that they considered both to be equivalent. To detect misuse, the experiment contained one pair of a very good and a very bad sorting. The decisions of users who preferred the bad sorting here were discarded. 
The number of different arrangements were 32 for the color set and 23 each for the three image sets, (giving 496 pairs for the color set and 253 pairs for each image set). Each pair was evaluated by at least 35 users. 

For each comparison of $S_i$ with $S_j$, the preferred arrangement gets one point. In case of a tie, both get half a point each. Let $v_r(S_i, S_j)$ be the points received by $S_i$ in the $r$\textsuperscript{th} out of $R$ comparisons between $S_i$ and $S_j$. Let 
\begin{equation}
    P(S_i, S_j) = \frac{1}{R}\sum_{r=1}^R v_r(S_i, S_j)
\end{equation}
be the probability that $S_i$ receives a higher quality assessment in comparison to $S_j$, ($P(S_i, S_j) + P(S_j, S_i) = 1$). The final user score for $S_i$ is defined by
\begin{equation}
    \score(S_i) = \sum_{j} P(S_i, S_j)
\end{equation}
Because the number of comparisons per pair was quite high and nearly constant, sophisticated methods for unbalanced pairwise comparison such as the Bradley-Terry model \cite{BradleyTerry, hunter_2004_mm} were not necessary since they provided an equal ranking.  

\begin{figure}[h!]
    \centering
    \includegraphics[width=0.7\linewidth]{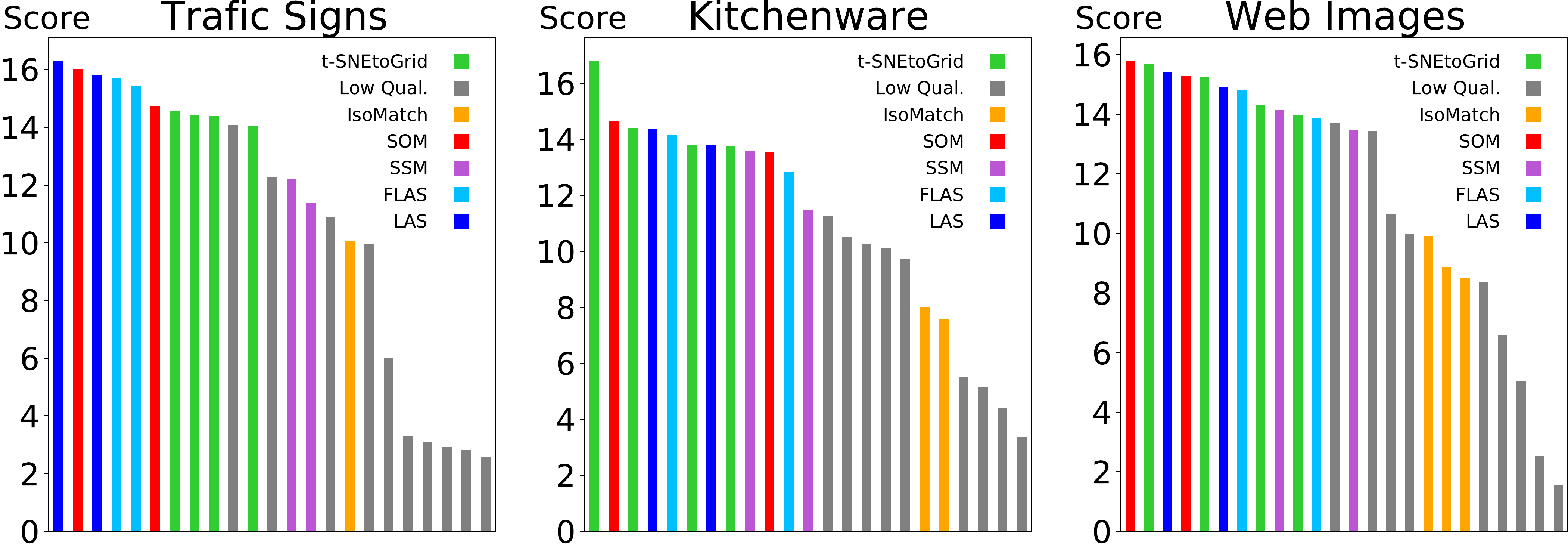}
    \caption{
            User scores for the image sets. The user scores for the color set can be seen in Figure \ref{fig:compare_experiment_result_colors}.} 
    \label{fig:user_scores}
\end{figure}

\begin{figure}[h!]
    \centering
    \includegraphics[width=0.6\linewidth]{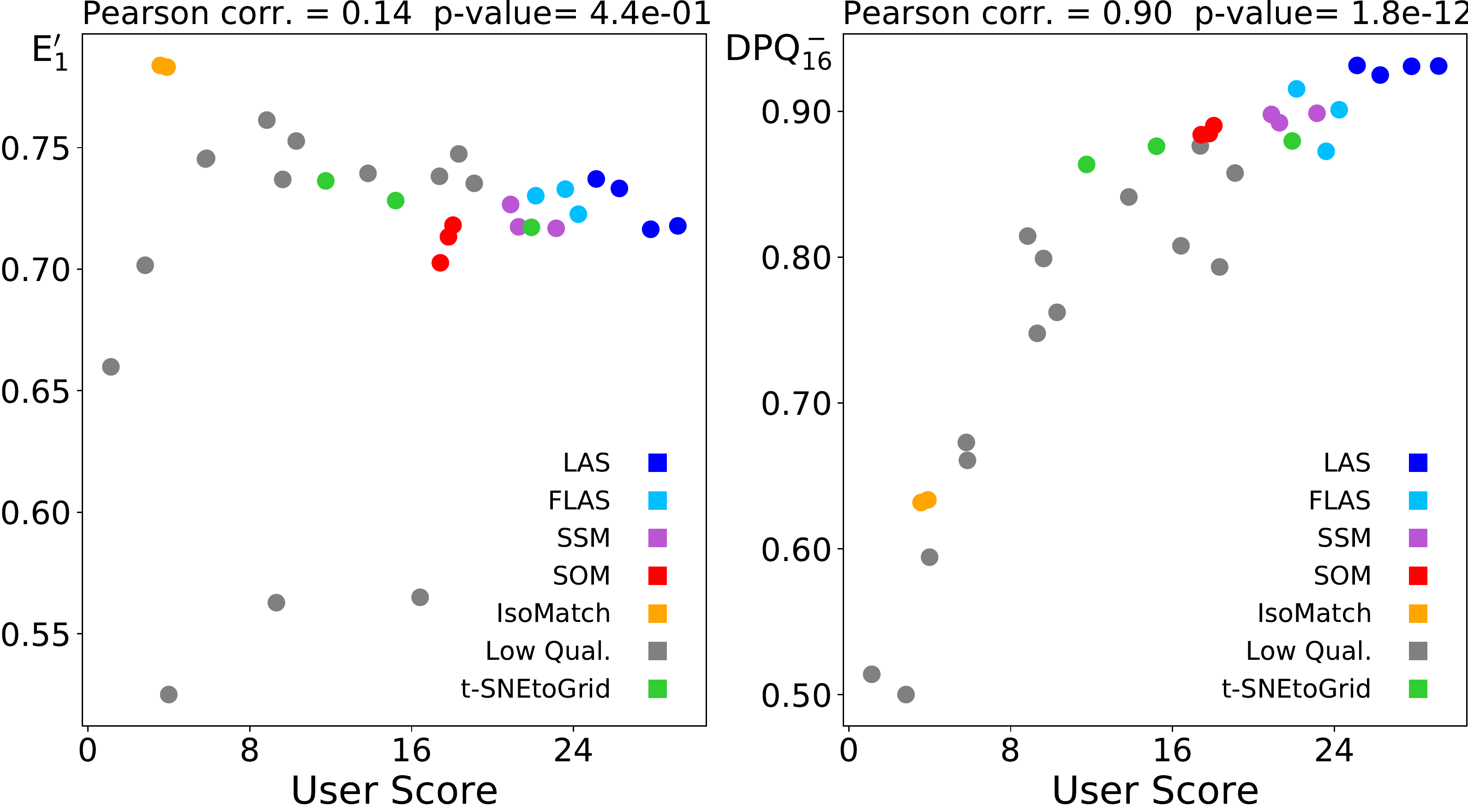}
    \caption{
            User scores correlated with energy function $\E_1'$ (left) and distance preservation quality $\DPQ_{16}^-$ (right) for the RGB color set.}
    \label{fig:compare_experiment_result_colors}
\end{figure}

\begin{figure}[!h]
    \centering
    \includegraphics[width=0.6\linewidth]{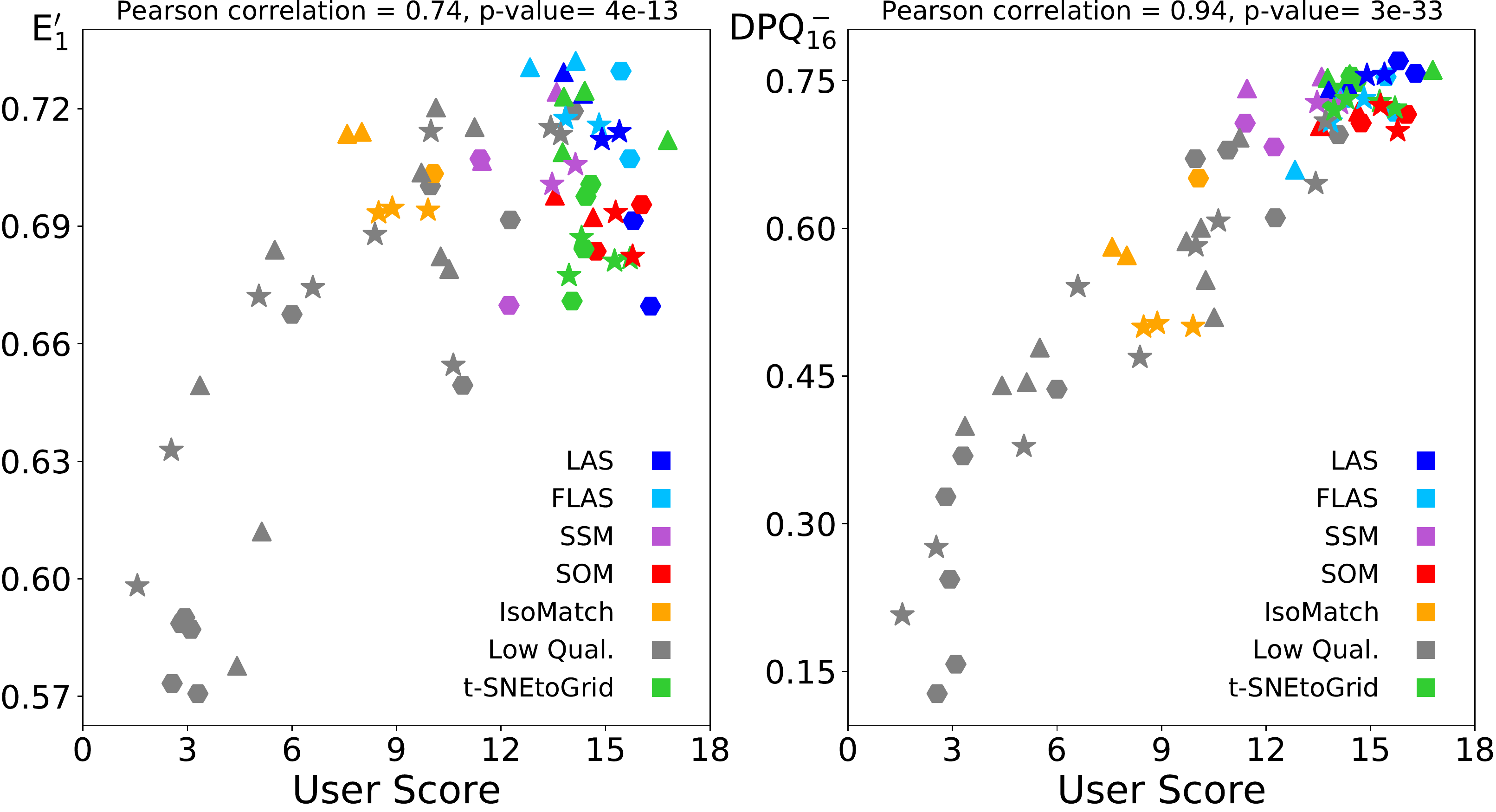}
    \caption{ User scores correlated with the energy function $\E_1'$ (left) and  $\DPQ_{16}^-$ (right) for the image sets: traffic signs (\tiny $\hexagon$\normalsize), kitchenware (\small{$\blacktriangle$}\normalsize), and web images (\tiny{$\bigstar$}\normalsize). The correlation between users scores and $DPQ_{16}^-$ is higher than that of $E_1'$.
    }
    \label{fig:compare_experiment_results_images}
\end{figure}

The overall result of the user evaluation of the arrangements is shown in Figure \ref{fig:user_scores}. Figures \ref{fig:compare_experiment_result_colors} and \ref{fig:compare_experiment_results_images} show the relationship between user ratings and the values of the $\E_1'$ and $\DPQ_{16}^-$ metrics for the color set and the three image sets.
It can be seen that the Pearson correlation is significantly higher for $\DPQ_{16}^-$ compared to $\E_1'$. In the case of RGB colors, users liked the LAS arrangements the best. For the image sets, there is no clear winning method. 
The \hbox{t-SNEtoGrid} method obtained rather low scores for the RGB colors, but much higher ones for the image sets.

\begin{figure}[h!]
     \centering
    \begin{subfigure}[t]{0.28\textwidth}
        {\includegraphics[width=\textwidth]{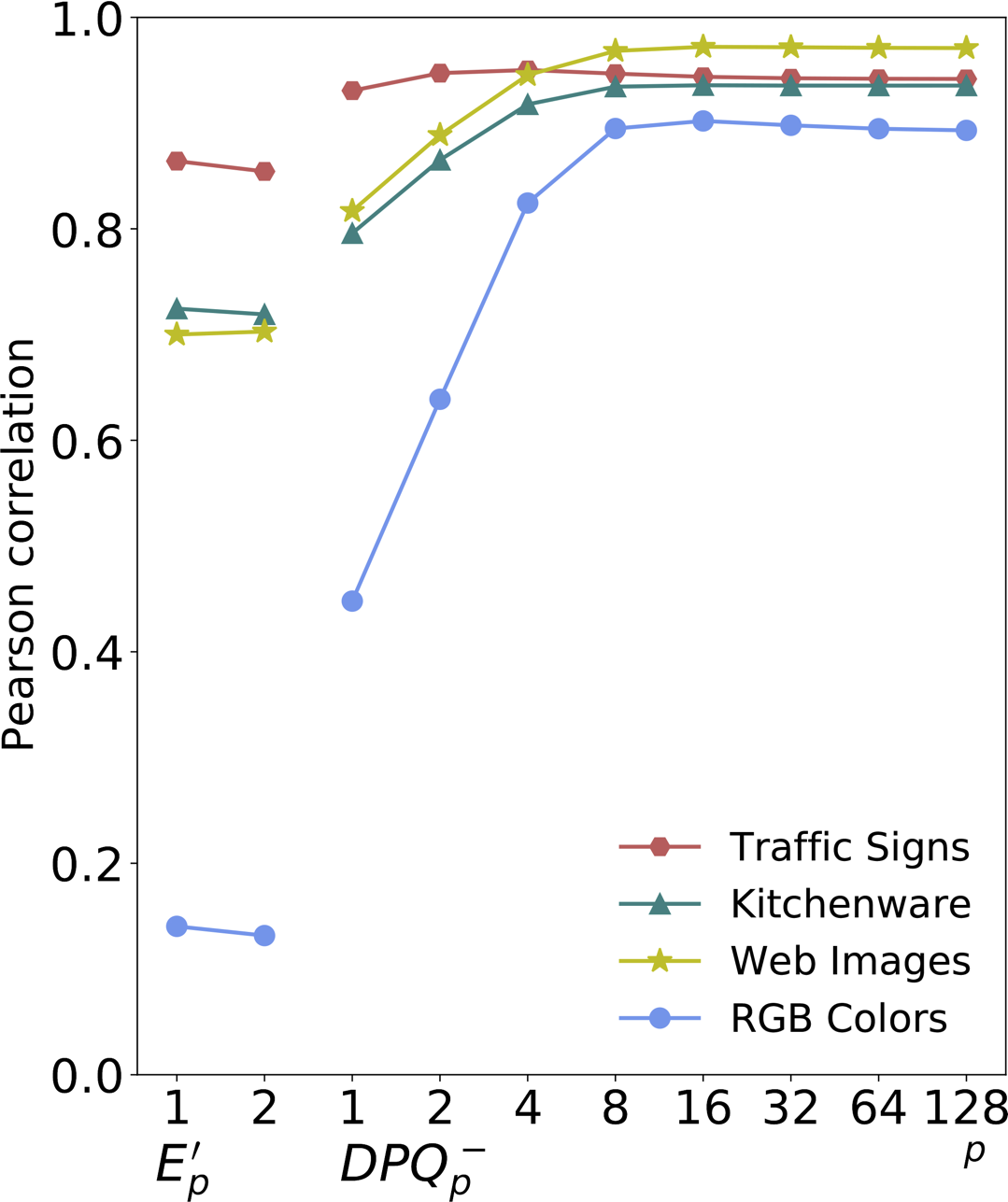}}
    \end{subfigure}
    \hspace{-0.0\linewidth}
    \begin{subfigure}[t]{0.28\textwidth}
        {\includegraphics[width=\textwidth]{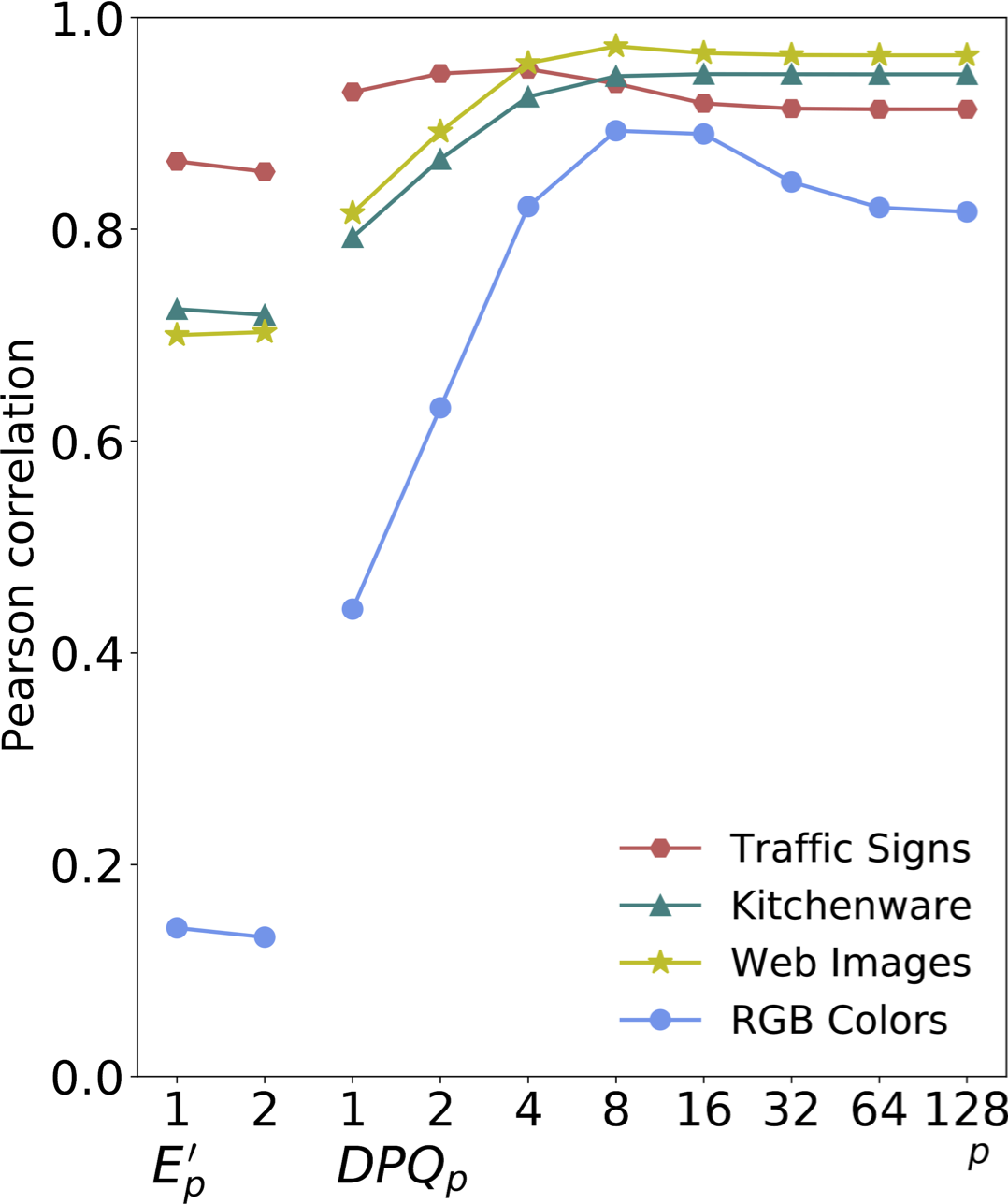}}
    \end{subfigure}
    \caption{The correlation of user scores with the \hbox{metrics} $\E_p'$ and $\DPQ_{p}$ with respect to the $p$ values for the color and image sets. For equal 2D distances the DPQ metric was computed using mean HD distances (left) and sorted HD distances (right) (see Figure \ref{fig:blau_rot_lila}).
 }\label{fig:compare_experiment_results_correlation}
\end{figure}

Figure \ref{fig:compare_experiment_results_correlation} shows the degree of correlation between user scores and quality metrics for different $p$ values of the $\E_p'$ and the $\DPQ_{p}$ metrics. 
For all four sets, the correlation of $\DPQ$ with the user scores is higher than that of $\E_1'$ and $\E_2'$ for all $p$ values. 
For predicting user scores, $DPQ_p^-$ values (using mean HD distances for equal 2D distances) with higher $p$ values give the best results (see left).
The high correlation for larger $p$ values with the user scores could indicate that users essentially pay attention to how well the immediate nearest neighbors have been preserved.

\subsection{Evaluation of User Search Time}
\label{sec:Evaluation of User Search Time}
\begin{figure}[!t]
    \centering
    \includegraphics[width=0.525\linewidth]{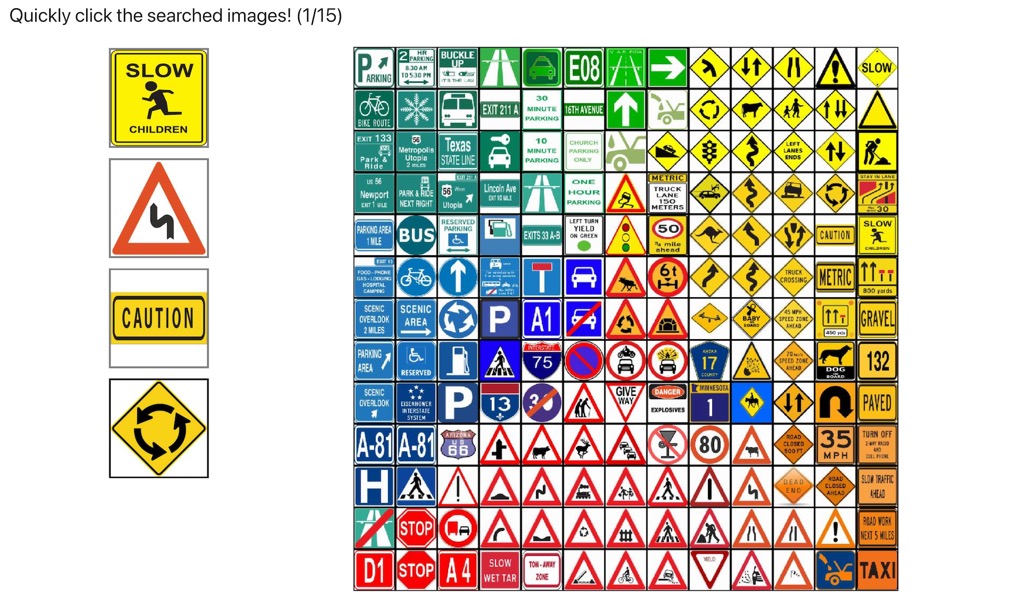}
    \caption{%
             Image search. In this study we ask users to quickly find the images shown on the left.
    }
    \label{fig:search_experiment}
\end{figure}

In the second part of the user study, the users were shown different arrangements in which they were asked to find four images in each case. 
The four images to be searched were randomly chosen and shown one after the other. As soon as one image was found, the next one was displayed. 
Participants were asked to pause only when they had found a group of four images, but not during a search. 
At the beginning, users were given a trial run to familiarize themselves with the task. 
Here, the time was not recorded. 
Figure \ref{fig:search_experiment} shows a screenshot of the search experiment. 
The overall set of arrangements was identical to those from the first part, in which the pairs had to be evaluated. 

Obviously, the task of finding specific images varies in difficulty depending on the image to be found. In addition, the participants are characterized by their varying search abilities. However, a total of more than 28000 search tasks were performed, each with four images to be found. This means that for each arrangement, more than 400 search tasks were performed for four images each. This compensated for differences in both the difficulty of the search and the abilities of the participants.

\begin{figure}[!h]
    \centering
    \includegraphics[width=0.6\linewidth]{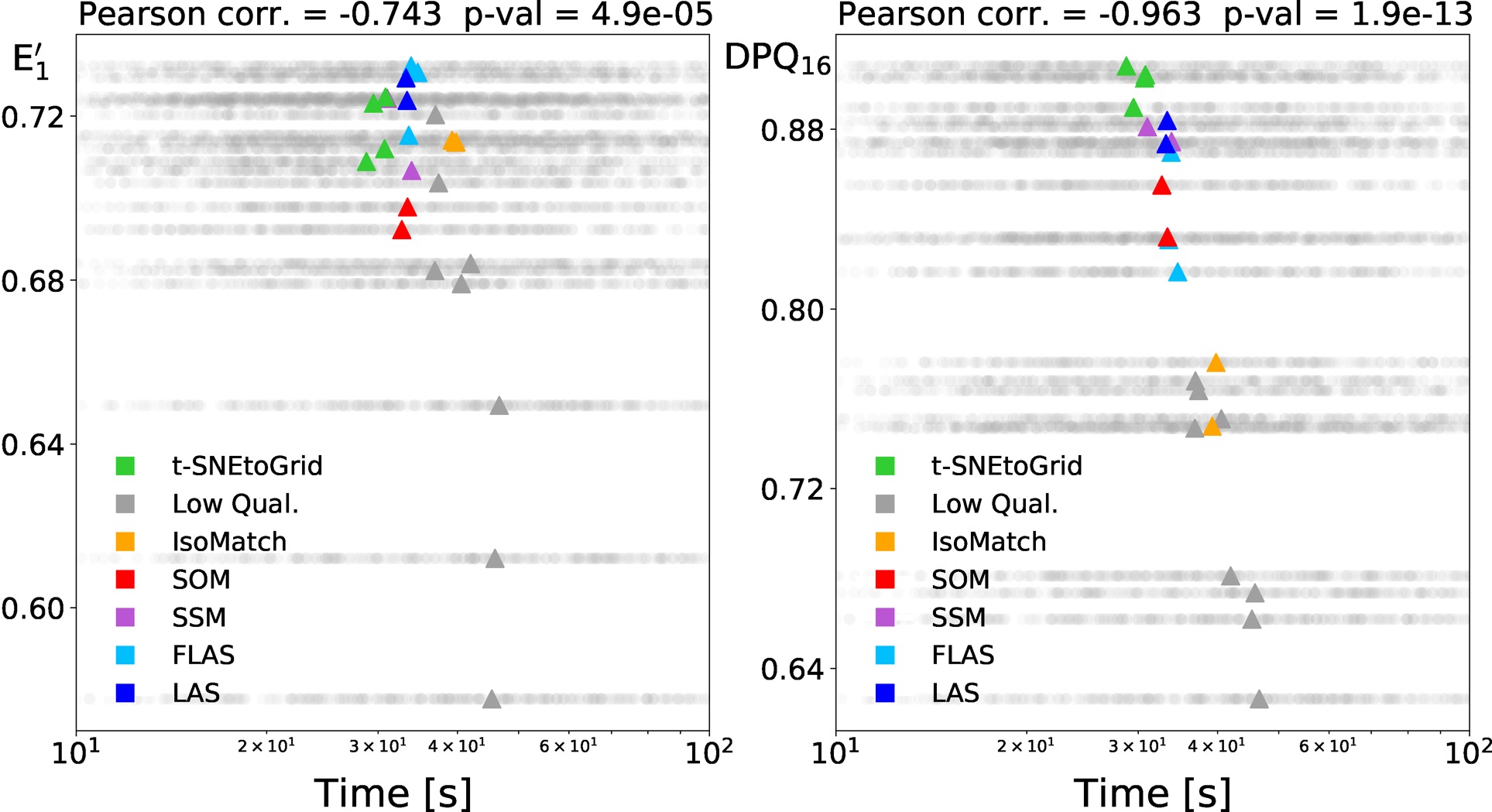}
    \caption{%
             Distribution of log search times versus metrics: $\E_1'$ (left) and $\DPQ_{16}$ (right) for the kitchenware image set. 
             The median search times are shown as colored dots.
    }
    \label{fig:search_experiment_kitchenware}
\end{figure}

The search times required for each of the 23 arrangements per image set were recorded. It was found that the time distribution of the searches is approximately log-normal. 
Search times that fell outside the upper three standard deviations were discarded to filter out experiments that were likely to have been interrupted. 
Figure \ref{fig:search_experiment_kitchenware} shows the distribution of the search times for the kitchenware image arrangements. The median values of the search times of the different arrangements are shown as colored dots. Again it can be seen that the correlation of the median search times is higher with the $\DPQ_{16}$ metric than with the $\E_1'$ metric.    

\begin{figure}[!h]
    \centering
    \includegraphics[width=0.7\linewidth]{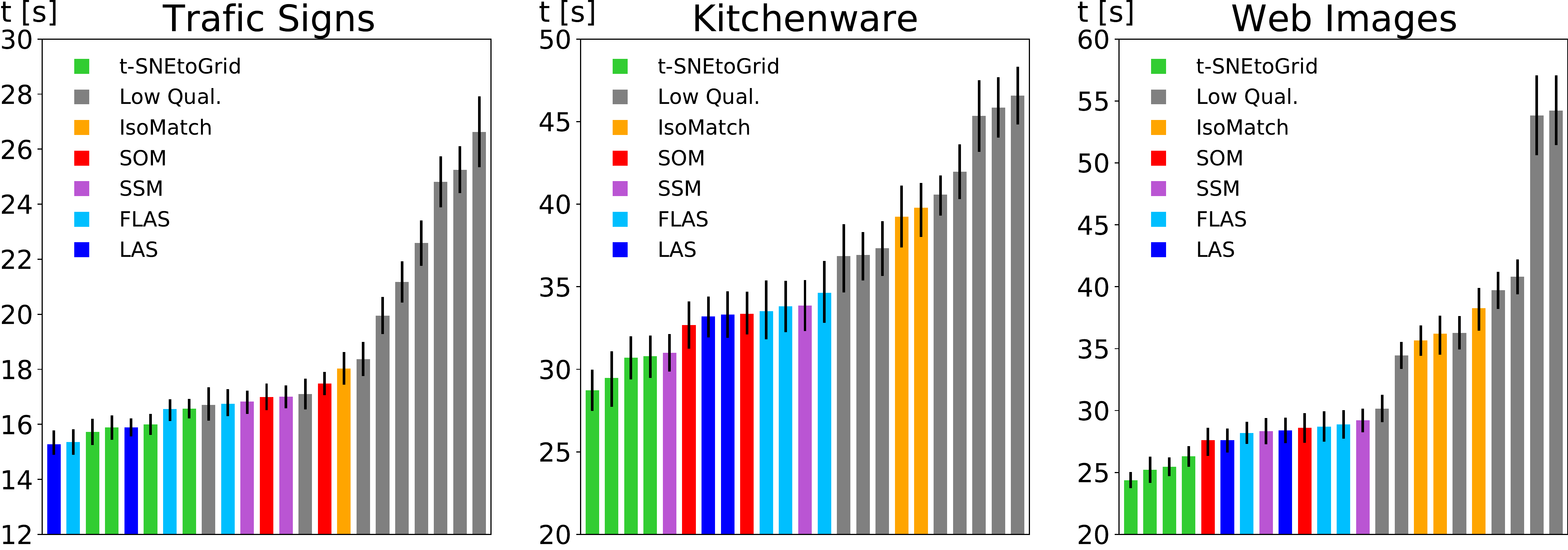}
    \caption{%
             The median search times for the three image sets. The standard errors of the medians are indicated. The time axis is clipped to better visualize differences of the median search times. 
    }
    \label{fig:search_times}
\end{figure}

Figure \ref{fig:search_times} shows the median values of the search times for all arrangements, sorted from the arrangement where the images were found the fastest to the one where the image search took the longest. The standard error of the median search times was determined by bootstrapping with 10000 runs. 
While the ranking order of the algorithms is similar to the order of the user preferences, it occasionally differs, suggesting that an apparently well-sorted arrangement is only conditionally indicative of finding images quickly. 

\begin{figure}[!h]
    \centering
    \includegraphics[width=0.7\linewidth]{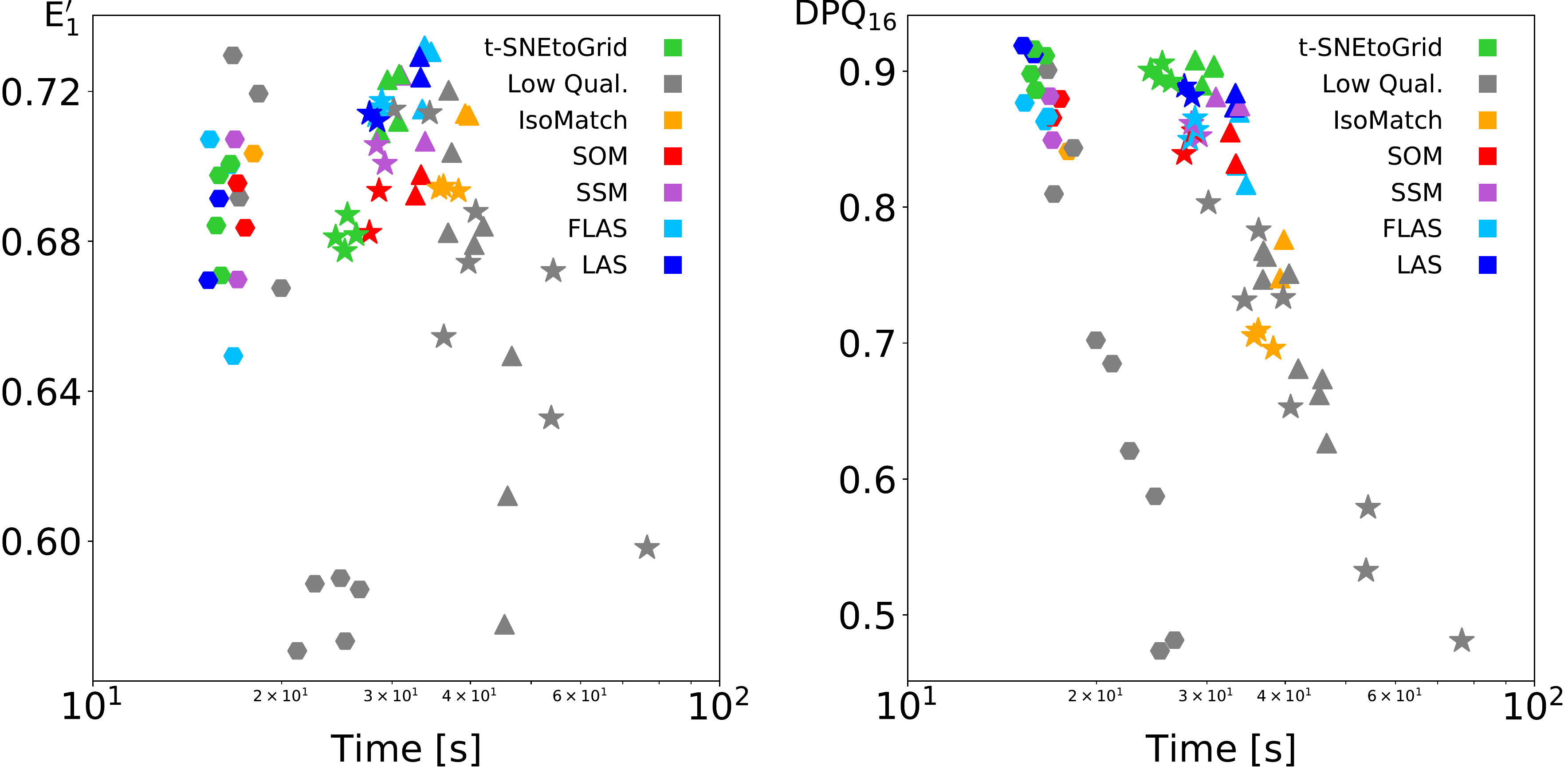}
    \caption{Median search time correlated with energy function $\E_1'$ (left) and distance preservation quality $\DPQ_{16}$ (right) for the image sets: traffic signs (\tiny $\hexagon$\normalsize), kitchenware (\small{$\blacktriangle$}\normalsize), and web images (\tiny{$\bigstar$}\normalsize).}
    \label{fig:median_search_times}
\end{figure}

Figure \ref{fig:median_search_times} compares the median search times with the metrics $\E_1'$ and $\DPQ_{16}$. Again, the (negative) correlation of $\DPQ$ with search time is much higher than that of the normalized energy function. 

\begin{figure}[h!]
     \centering
    \begin{subfigure}[t]{0.28\textwidth}
        {\includegraphics[width=\textwidth]{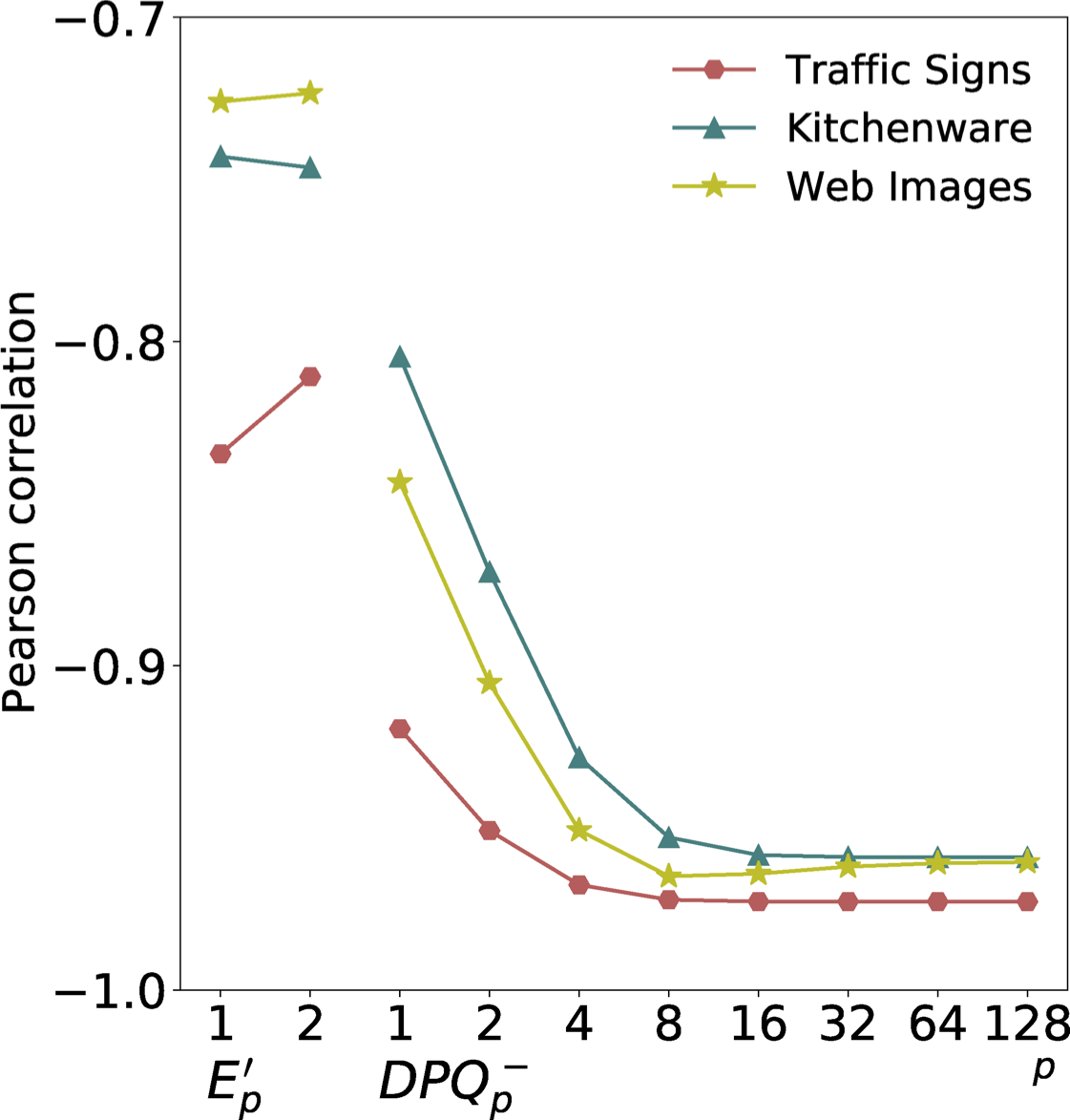}}
    \end{subfigure}
    \begin{subfigure}[t]{0.28\textwidth}
        {\includegraphics[width=\textwidth]{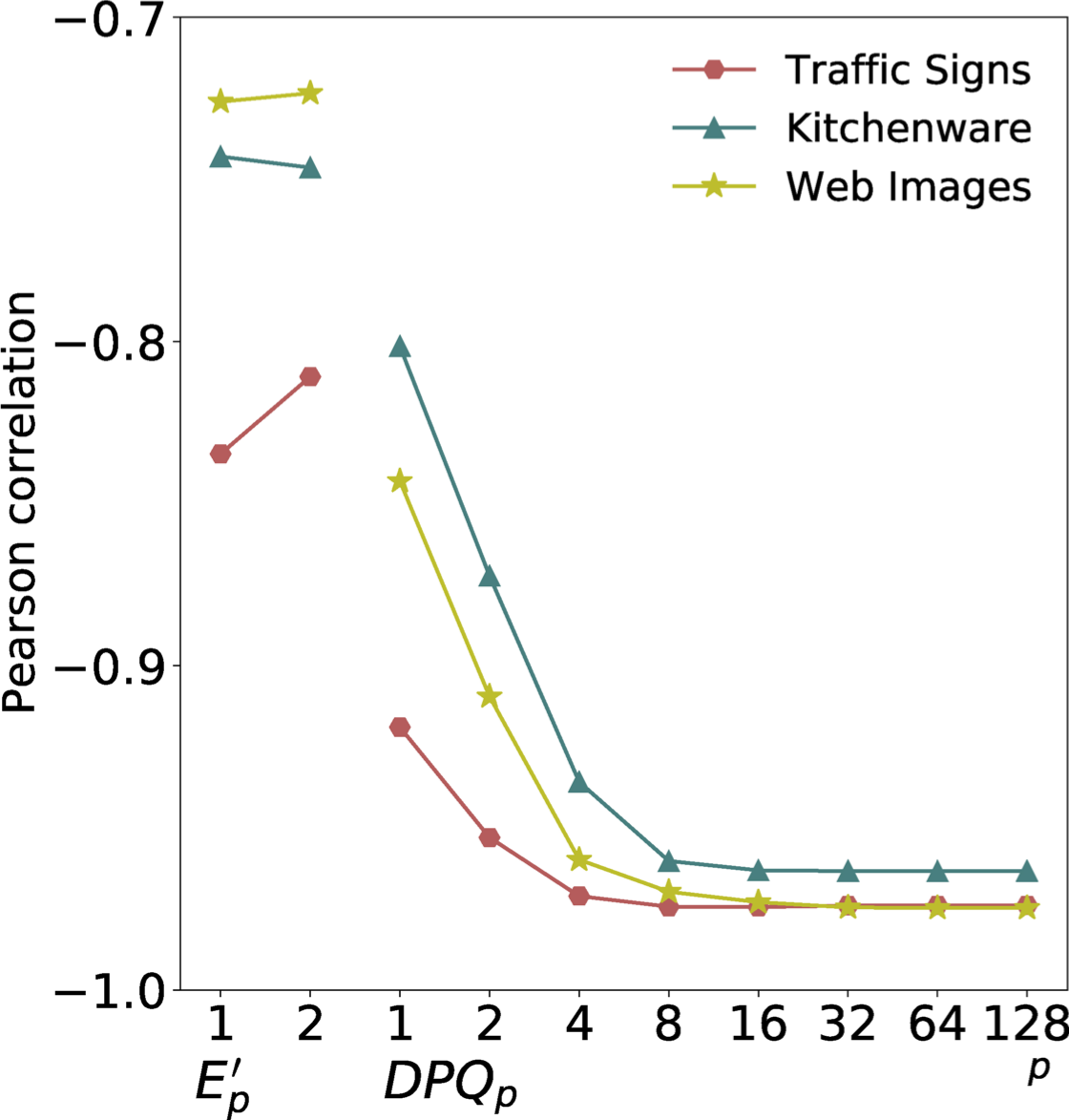}}
    \end{subfigure}
    \caption{The correlation of median image search time with the \hbox{metrics} $\E_p'$ and $\DPQ_{p}^-$ / $\DPQ_{p}$ with respect to the $p$ values.
 }\label{fig:search_experiment_results_correlation}
\end{figure}

To evaluate the degree of correlation, Figure \ref{fig:search_experiment_results_correlation} compares the normalized energy function ($E_1'$ and $E_2'$) with the distance preservation quality ($\DPQ_q$ and $\DPQ_q^-$). 
Again it can be seen that $\DPQ_p$ outperforms $E_p$. 
Contrary to the user preference evaluation, for image retrieval tasks $\DPQ_q$ performs slightly better than $\DPQ_q^-$. 
Both show maximum correlation for high $p$ values. This in turn indicates that it seems to be most important for people to locate similar images very close to each other in order to find them quickly, as hypothesized in Figure \ref{fig:blau_rot_lila}.

We also investigated the use of the squared $L2$ distance instead of the $L2$ distance to calculate the $\DPQ$, the correlations remained similar, but were slightly lower. 
Other distance transformations could be investigated, but this is beyond the scope of this paper.



\section{Qualitative and Quantitative
Comparisons}

\subsection{Quality and Run Time Comparison}

To get a better understanding about the behaviour of FLAS and other 2D grid-arranging algorithms using different hyperparameter settings, we conducted a series of experiments. 
Since the run time strongly depends on the hardware and implementation quality, the numbers given in this section only serve as comparative values.
In the previous section, $\DPQ_{16}$ has shown high correlation with user preferences and performance, we therefore use it when comparing algorithms in terms of their achieved "quality" and the run time required to generate the sorted arrangement.

Our test machine is a Ryzen 2700x CPU with a fixed core clock of 4.0 GHz and 64GB of DDR4 RAM running at 2133 MHz.  
The tested algorithms were all implemented in Java and executed with the JRE $1.8.0\_321$ on Windows 10. Only the single threaded sorting time was measured.
As much code as possible was re-used (e.g. the solver of LAS, FLAS, IsoMatch, and t-SNEtoGrid) to make the comparison as consistent as possible. 

The IsoMap and t-SNE projection implementation is from the popular library SMILE \cite{Li2014Smile} (version 2.6). 
The SSM code is an implementation adapted from \cite{Strong2014} to match the characteristics of our implementation of SOM, LAS, and FLAS. 
At startup, all data is loaded into memory. Then the averaged run time and $\DPQ_{16}$ value of 100 runs were recorded. 
We ensured the algorithms received the same initial order of images for all the runs.

\begin{figure}[h!]
    \centering
    \subfloat {
        \includegraphics[width=0.48\linewidth]{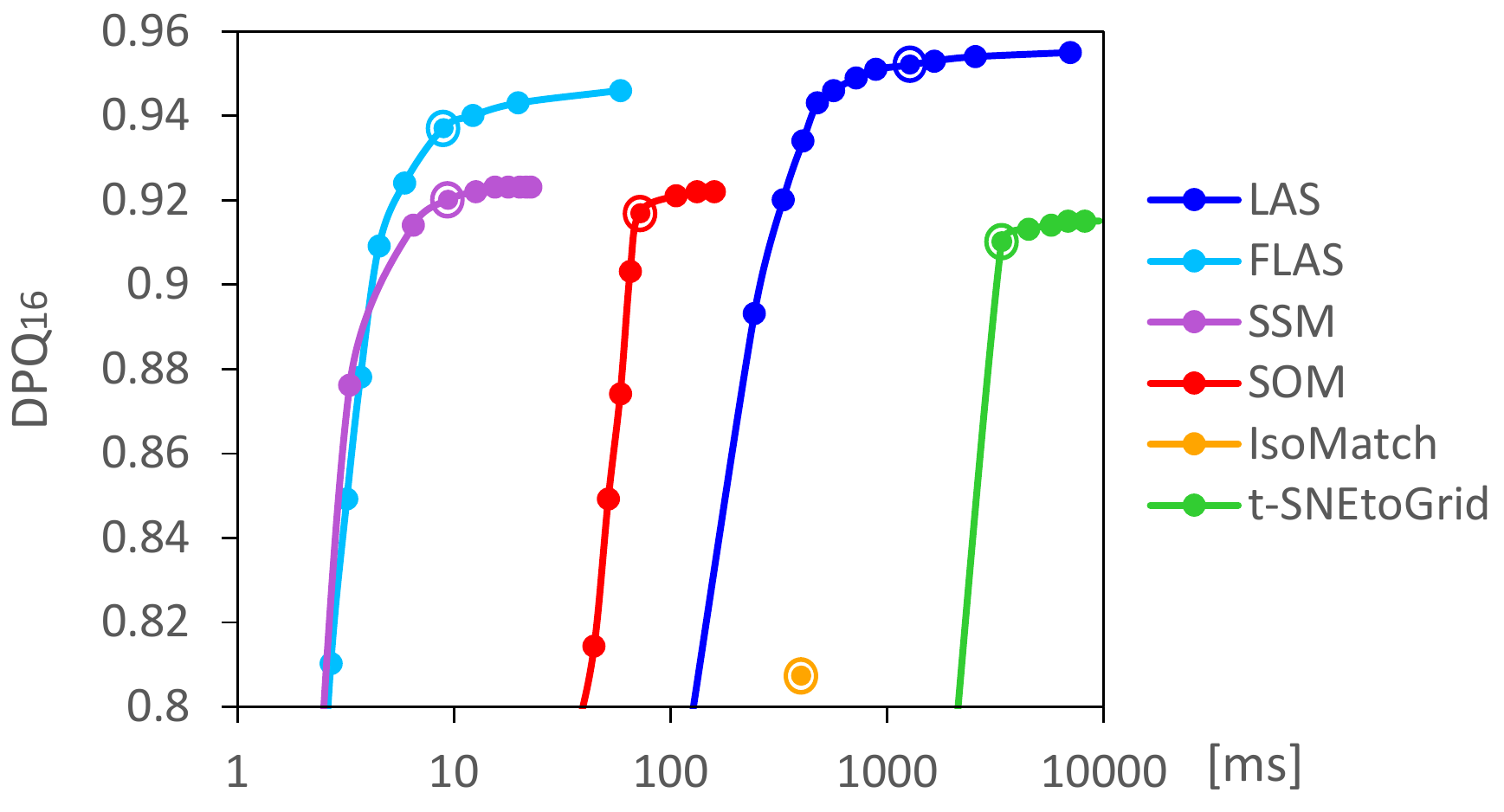}
    }
    \subfloat {
        \includegraphics[width=0.48\linewidth]{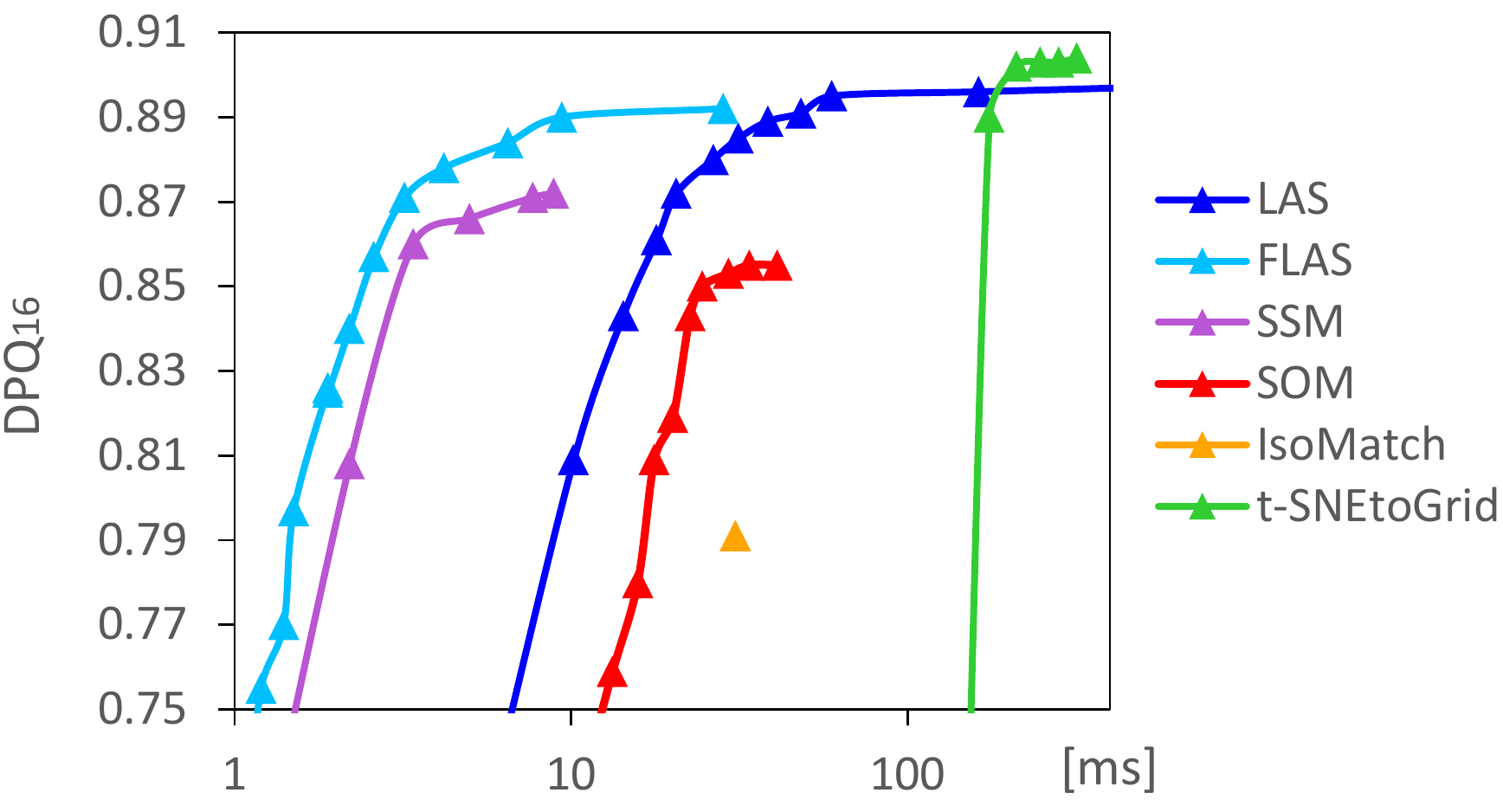}
    }
    \caption{Mean $\DPQ$ vs. mean run time using different parameter settings for the 1024 RGB colors (left) and the 256 kitchenware images (right). ${\circledot}$ indicates parameters used in Figure \ref{fig:algorithm_scalability}.}
    \label{fig:algorithm_param_runtime_comparison}
\end{figure}

There are different hyperparameters that can be tuned. Some of them affect the run time and/or the quality of the arrangement, while others result in only minor changes. 
Figure \ref{fig:algorithm_param_runtime_comparison} shows the relationship between speed and quality when varying the hyperparameters.
For t-SNE, SSM and SOM the number of iterations were changed, the t-SNE  learning rate (eta) of was set to 200. 
For LAS and FLAS the radius reduction factor was gradually reduced from 0.99 to 0, while the initial radius factor was 0.35 and 0.5 respectively. FLAS used 9 swap candidates per iteration. 
IsoMatch has only the k-neighbor setting which does not influence the quality nor the run time and therefore produces only a single data point in the plots. 
For small data sets like the 256 kitchenware images, FLAS offers the best trade-off between ($\DPQ$) and computation time. 
LAS and t-SNE can produce higher $\DPQ_{16}$ values but are 10-100 times slower.
There is no reason to use a SSM or SOM, since both are either slower or generate inferior arrangements. 
For the 1024 random RGB colors, LAS and FLAS yielded the highest distance preservation quality.

\begin{figure}[!htb]
  \centering
  \mbox{}
  \includegraphics[width=0.5\linewidth]{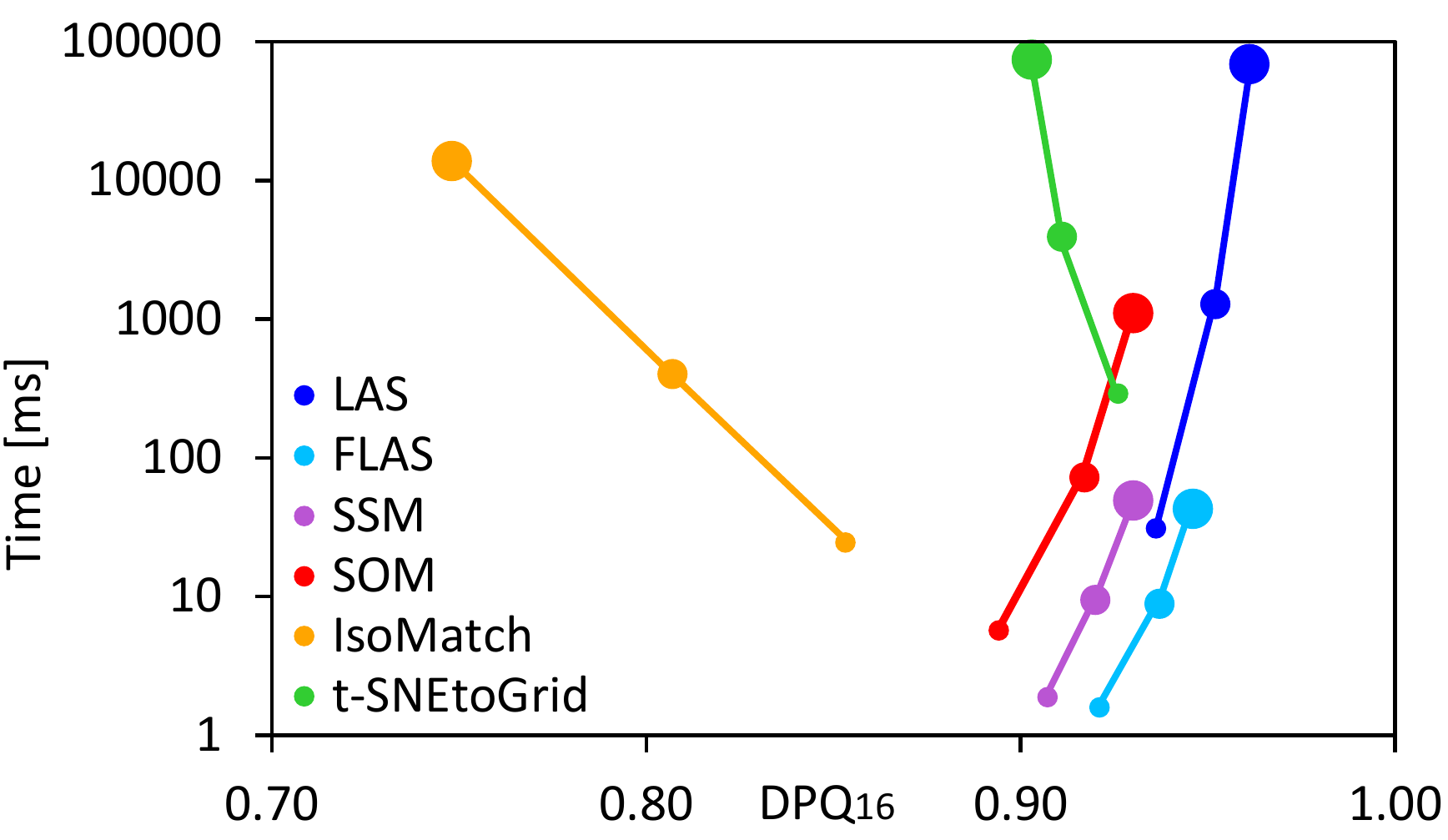}
  \caption{Scalability of different algorithms. The mean $\DPQ$ and mean computation time for data sets of 256 ({\smalldot}), 1024 ({\mediumdot}), and 4096 ({\largedot}) random colors using fixed hyperparameters. 
}
\label{fig:algorithm_scalability}
\end{figure}

\begin{figure*}[t!]
    \captionsetup{singlelinecheck = false} 
    \begin{subfigure}[t]{0.155\textwidth}
        \raisebox{-\height}{\includegraphics[width=\textwidth]{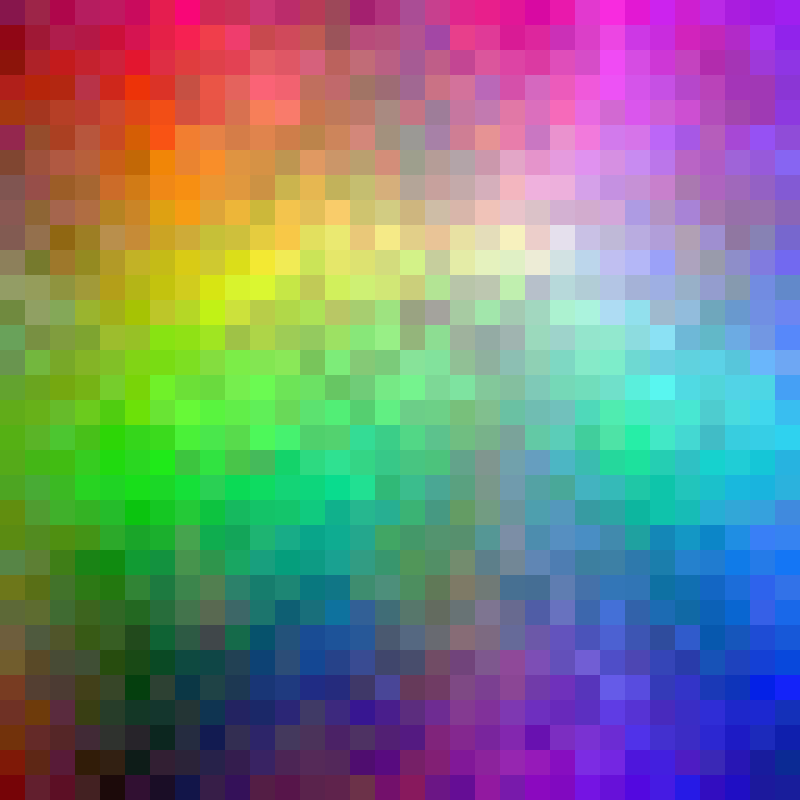}}
        \captionsetup{font=scriptsize}
        \caption{LAS \\${\DPQ_{16}}$: 0.95 \enspace$\E'_1$: 0.72 }
    \end{subfigure}
    \hfill
    \begin{subfigure}[t]{0.155\textwidth}
        \raisebox{-\height}{\includegraphics[width=\textwidth]{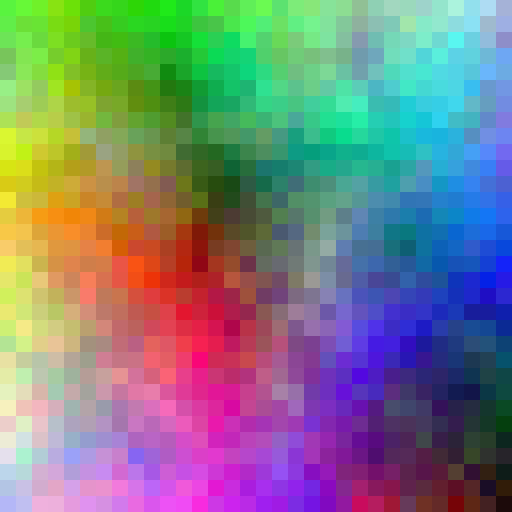}}
        \captionsetup{font=scriptsize}
        \caption{FLAS \\${\DPQ_{16}}$: 0.94 \enspace$\E'_1$: 0.72 }
    \end{subfigure}
    \hfill
    \begin{subfigure}[t]{0.155\textwidth}
        \raisebox{-\height}{\includegraphics[width=\textwidth]{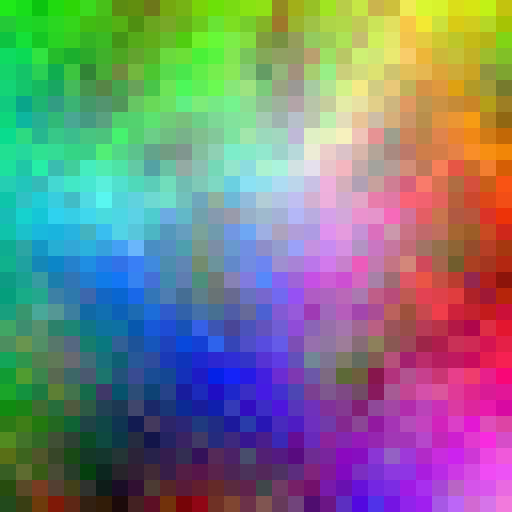}}
        \captionsetup{font=scriptsize}
        \caption{SSM \\${\DPQ_{16}}$: 0.92 \enspace$\E'_1$: 0.67 }
    \end{subfigure}
    \hfill
    \begin{subfigure}[t]{0.155\textwidth}
        \raisebox{-\height}{\includegraphics[width=\textwidth]{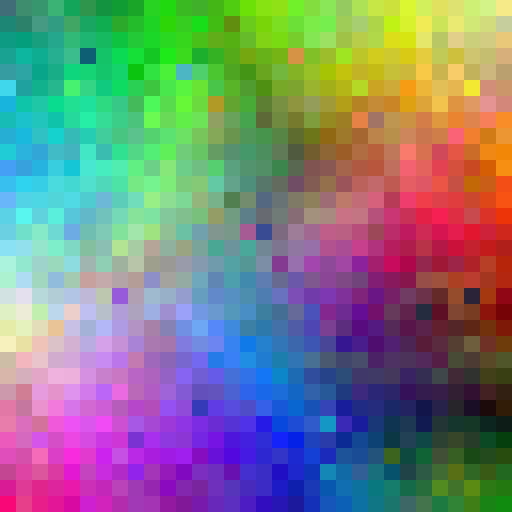}}
        \captionsetup{font=scriptsize}
        \caption{SOM \\${\DPQ_{16}}$: 0.92 \enspace$\E'_1$: 0.72 }
    \end{subfigure}
    \hfill
    \begin{subfigure}[t]{0.155\textwidth}
        \raisebox{-\height}{\includegraphics[width=\textwidth]{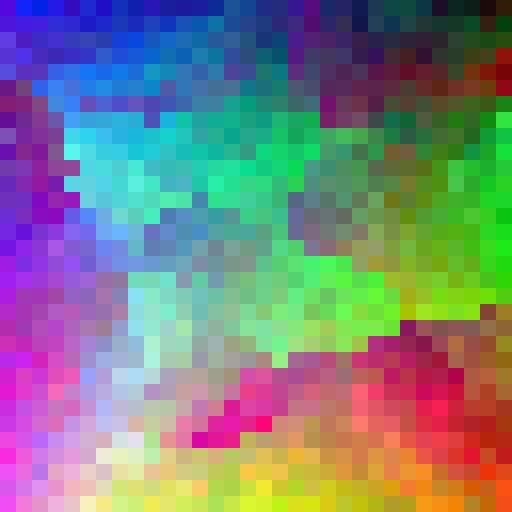}}
        \captionsetup{font=scriptsize}
        \caption{t-SNEtoGrid \\${\DPQ_{16}}$: 0.91 \enspace$\E'_1$: 0.75 }
    \end{subfigure}
    \hfill
    \begin{subfigure}[t]{0.155\textwidth}
        \raisebox{-\height}{\includegraphics[width=\textwidth]{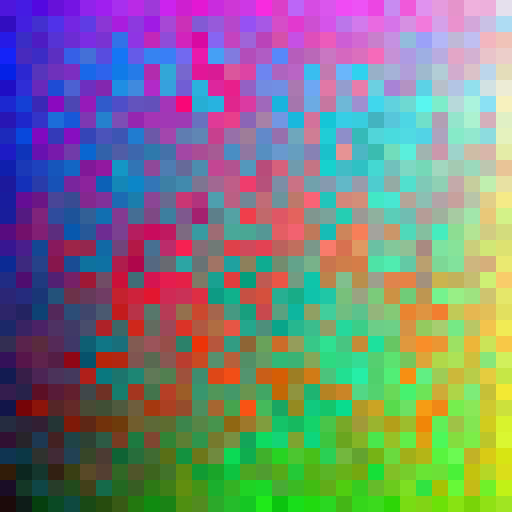}}
        \captionsetup{font=scriptsize}
        \caption{IsoMatch \\${\DPQ_{16}}$: 0.81 \enspace$\E'_1$: 0.76 }
    \end{subfigure}
       \captionsetup{singlelinecheck = false} 
    \begin{subfigure}[t]{0.155\textwidth}
        \raisebox{-\height}{\includegraphics[width=\textwidth]{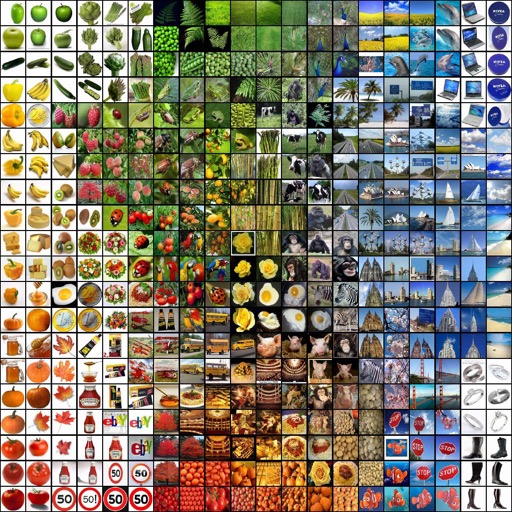}}
        \captionsetup{font=scriptsize}
        \caption{t-SNEtoGrid, U. sc.: 15.3\\${\DPQ_{16}}$: 0.90 \enspace$\E'_1$: 0.68 }
    \end{subfigure}
    \hfill
    \begin{subfigure}[t]{0.155\textwidth}
        \raisebox{-\height}{\includegraphics[width=\textwidth]{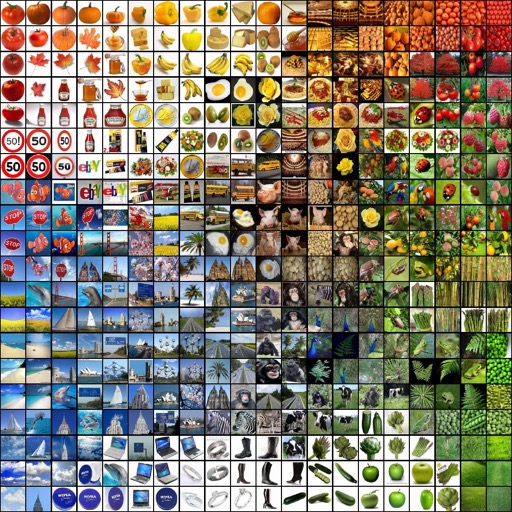}}
        \captionsetup{font=scriptsize}
        \caption{LAS, User score: 15.4 \\${\DPQ_{16}}$: 0.89 \enspace$\E'_1$: 0.71 }
    \end{subfigure}
    \hfill
    \begin{subfigure}[t]{0.155\textwidth}
        \raisebox{-\height}{\includegraphics[width=\textwidth]{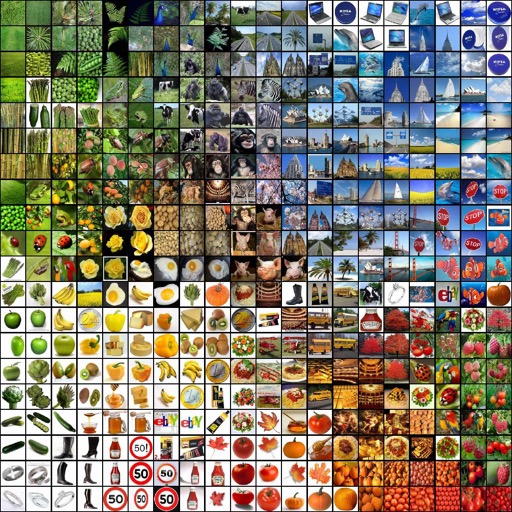}}
        \captionsetup{font=scriptsize}
        \caption{SOM, User score: 15.8 \\${\DPQ_{16}}$: 0.84 \enspace$\E'_1$: 0.68 }
    \end{subfigure}
    \hfill
    \begin{subfigure}[t]{0.155\textwidth}
        \raisebox{-\height}{\includegraphics[width=\textwidth]{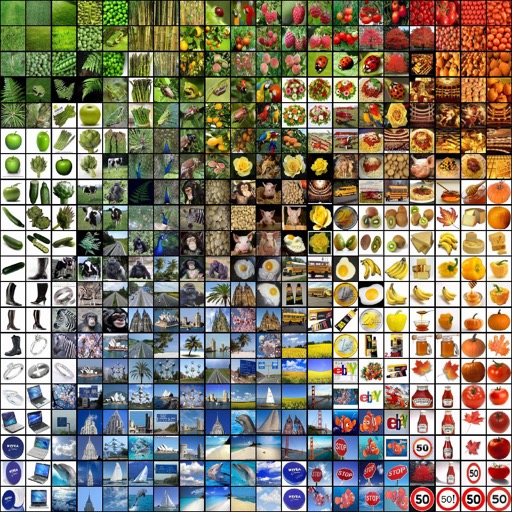}}
        \captionsetup{font=scriptsize}
        \caption{FLAS, User score: 13.7 \\${\DPQ_{16}}$: 0.85 \enspace$\E'_1$: 0.71 }
    \end{subfigure}
    \hfill
    \begin{subfigure}[t]{0.155\textwidth}
        \raisebox{-\height}{\includegraphics[width=\textwidth]{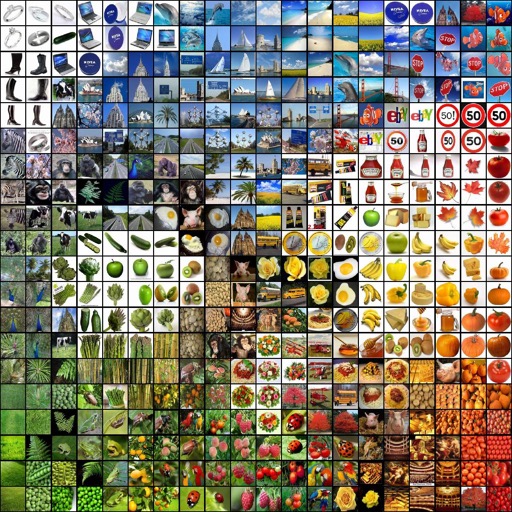}}
        \captionsetup{font=scriptsize}
        \caption{SSM, User score: 14.1 \\${\DPQ_{16}}$: 0.86 \enspace$\E'_1$: 0.71 }
    \end{subfigure}
    \hfill
    \begin{subfigure}[t]{0.155\textwidth}
        \raisebox{-\height}{\includegraphics[width=\textwidth]{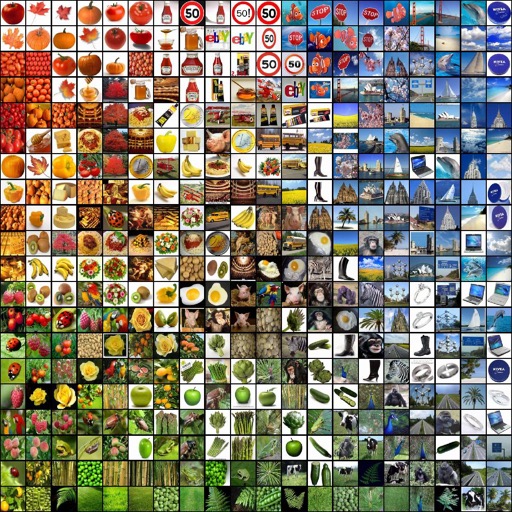}}
        \captionsetup{font=scriptsize}
        \caption{IsoMatch, User Score: 9.9 \\${\DPQ_{16}}$: 0.71 \enspace$\E'_1$: 0.69 }
    \end{subfigure}
    \caption{Comparison of different image sorting schemes. Top: Arrangements of the 1024 random RGB colors ordered by distance preservation quality. Bottom: Arrangements of the web image set ordered by the median user search time (fastest on the left). For each algorithm, the arrangement that provided the fastest search is shown (see Figure \ref{fig:median_search_times}).}
    \label{fig:color1024_comparison}
\end{figure*}

In order to compare the scalability of the analyzed algorithms, three data sets of different sizes were analyzed, containing 256, 1024, and 4096 random RGB colors. 
The hyperparameters of the points marked with a ${\circledot}$ in Figure \ref{fig:algorithm_param_runtime_comparison} were used for all the tests shown in Figure \ref{fig:algorithm_scalability}. 
It can be seen FLAS and SSM have the same scaling characteristics, while FLAS delivers better qualities. Even higher $\DPQ$ values can be achieved by LAS at the expense of run time. 
As the number of colors increases, arrangements with smoother gradients become possible, resulting in better quality. 
Most approaches can exploit this property, except for IsoMatch and t-SNEtoGrid. 
Both initially project to 2D and rely on a solver to map the overlapping and cluttered data points to the grid layout. 
Since the number of grid cells is equal to the number of data points, it is difficult for the solver to find a good mapping. This often results in hard edges, as can be seen in Figure \ref{fig:color1024_comparison}.

To summarize, LAS can be used for high quality arrangements, whereas FLAS should be used if the number of images is very high or fast execution is important.

\subsection{Visual Comparison}
The quantitative qualities measured for the different algorithms in the previous section are quite similar. 
However, a visual inspection reveals specific differences.  

For the 1024 RGB color data set, the run with the result closest to the $\DPQ_{16}$-mean was selected from the 100 test runs used for Figure \ref{fig:algorithm_scalability}. The corresponding arrangements are shown in the order of their distance preservation quality $\DPQ_{16}$ in the top of Figure \ref{fig:color1024_comparison}.
In addition, the normalized energy function value ($\E'_1$) is given. 
LAS has the smoothest overall arrangement, followed by FLAS and SSM. 
The SOM arrangement is disturbed by isolated, poorly
positioned colors, while the t-SNEtoGrid approach shows boundaries between regions.
This is due to previously separate groups of projected vectors being redistributed over the grid, resulting in visible boundaries where these regions touch. 
The noisy looking arrangement of IsoMatch is caused by the normalized energy function trying to equally preserve all distances. This leads to a kind of dithering of the vectors. 

Most of these effects are less visible when real images are used instead of colors. 
The lower row of Figure \ref{fig:color1024_comparison} shows the arrangements from our user study \ref{sec:Evaluation of User Search Time}, which required searching images in the web images set.  
The $\DPQ_{16}$ values, the $\E'_1$ values, and the user scores are given. The arrangements are ordered by the median time it took users to find the images they were looking for (from fastest to slowest).
t-SNEtoGrid again shows some boundaries between regions, but this time boundaries apparently help to better identify the individual image groups, reducing the time needed to find the images. 
The LAS, FLAS, SSM and SOM arrangements have a similar appearance for this data set. The dithered appearance of the IsoMatch arrangement apparently makes it harder for the users to find the searched images quickly.

\section{Applications}
In this section we consider a variety of applications of our algorithms introduced in Section \ref{Linear Assignment Sorting}.  

\subsection{Image Management Systems}
\begin{figure}[!h]
    \centering
    \includegraphics[height=0.35\linewidth ]{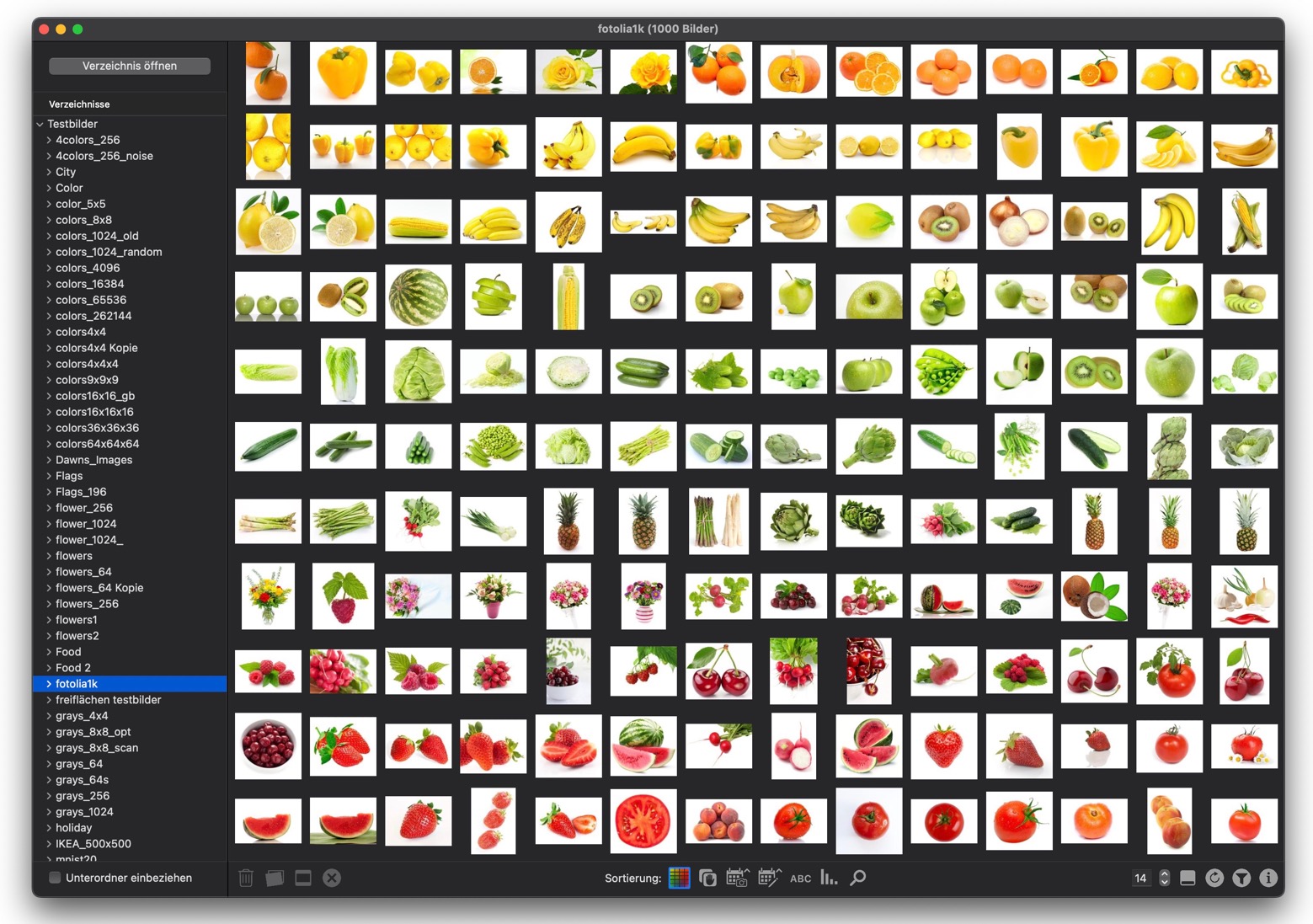}
    \caption{%
           When scrolling vertically for image management, images on horizontal lines should be similar to each other. 
    }
    \label{fig:picarrange}
\end{figure}

For browsing local images on a computer, a visually sorted display of the images can help to view more images at once. Since the user of a view tends to scroll in the vertical direction, it is important that the images in the horizontal row are similar to each other and one perceives clear changes in the vertical direction. This can be achieved by using a larger filter radius for horizontal filtering (see Figure \ref{fig:picarrange} showing the PicArrange app \cite{PicArrange} as an example).

\subsection{Image Exploration}
\begin{figure}[ht]
    \centering
        {\includegraphics[width=0.3\textwidth]{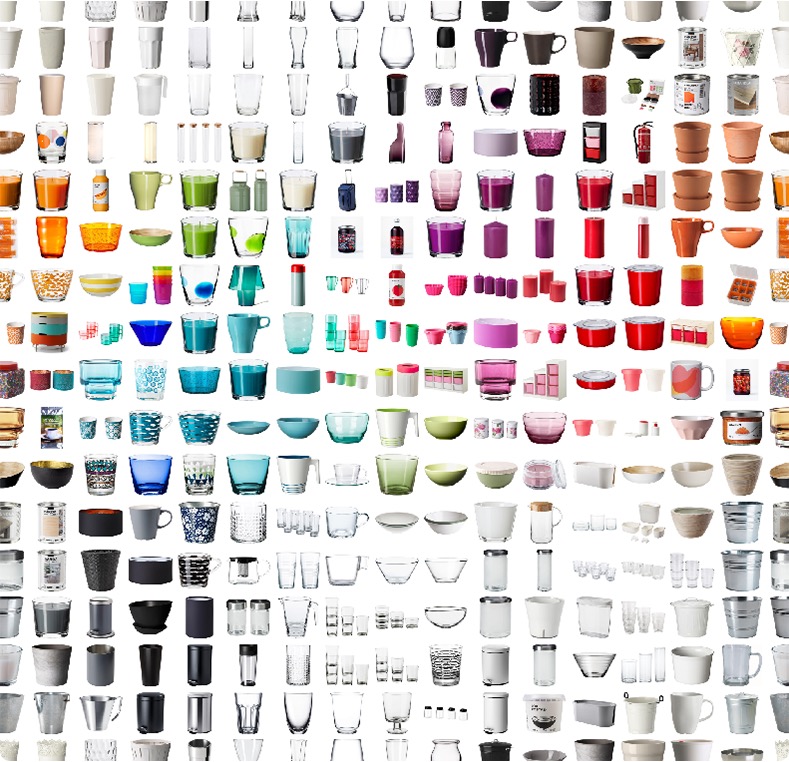}}
    \hspace{1cm}
        {\includegraphics[width=0.3\textwidth]{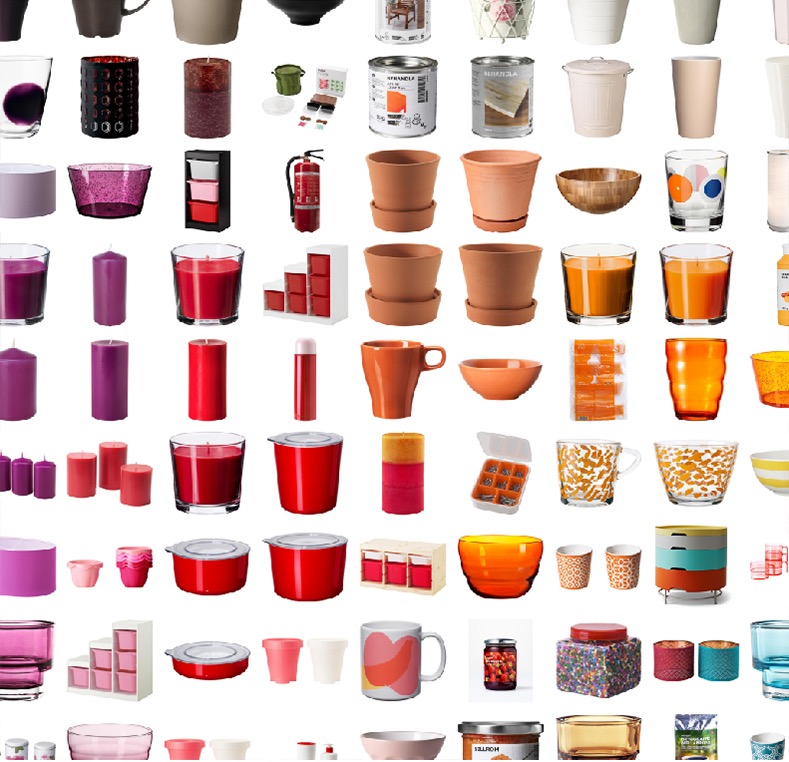}}
    \caption{Left: the entire collection wrapped, right: zoomed and dragged to center on orange pots
 }\label{fig:wrap}
\end{figure}

For very large sorted sets with millions of images, it may be useful to use a torus-shaped map which gives the impression of an endless plane. 
If one can navigate this plane in all directions it is possible to bring regions of interest into view. 
Such an arrangement can be achieved by using a wrapped filter operation. 
This means part of the low-pass filter kernel uses vectors from the opposite edge of the map. Figure \ref{fig:wrap} shows a (small) example of such a torus-shaped arrangement. 
If zooming is possible, images of interest can be found and inspected very easily.
This idea can be combined with a hierarchical pyramid of sorted maps that allows visual exploration for huge image sets. The user can explore the image pyramid with an interface similar to the mapping service (e.g. Google Maps). By dragging or zooming the map, other parts of the pyramid can be explored. The online tool \emph{wikiview.net} \cite{Wikiview} allows the exploration of millions of Wikimedia images using this approach. 

\subsection{Layouts with Special Constraints }
\begin{figure}[htb]
  \centering
  $\vcenter{\hbox{\includegraphics[width=.14\linewidth]{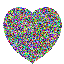}}}$
  \hspace{0.1\linewidth}
  $\vcenter{\hbox{\includegraphics[width=.14\linewidth]{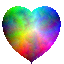}}}$
 
  \caption{2403 random RGB colors in a heart shape. Unsorted on the left and sorted with Linear Assignment Sorting on the right.
  }
  \label{fig:nonSwapable}
\end{figure}

Although our algorithms work with rectangular grids, other shapes can also be sorted. The map has the size of the rectangular bounding box of the desired shape, however, only the map positions within the shape can be assigned.
Figure \ref{fig:nonSwapable} shows an example where the colors to be sorted were only allowed within the shape of a heart.
The corresponding algorithm remains the same, the only difference is that the vectors of the assigned map positions are used to fill the rest of the the map positions. Each unassigned map position is filled with the nearest assigned vector of the map's constrained positions.

\begin{figure}[h]
 \centering
      \begin{subfigure}{0.49\linewidth}
        \centering
        \includegraphics[width=\linewidth]{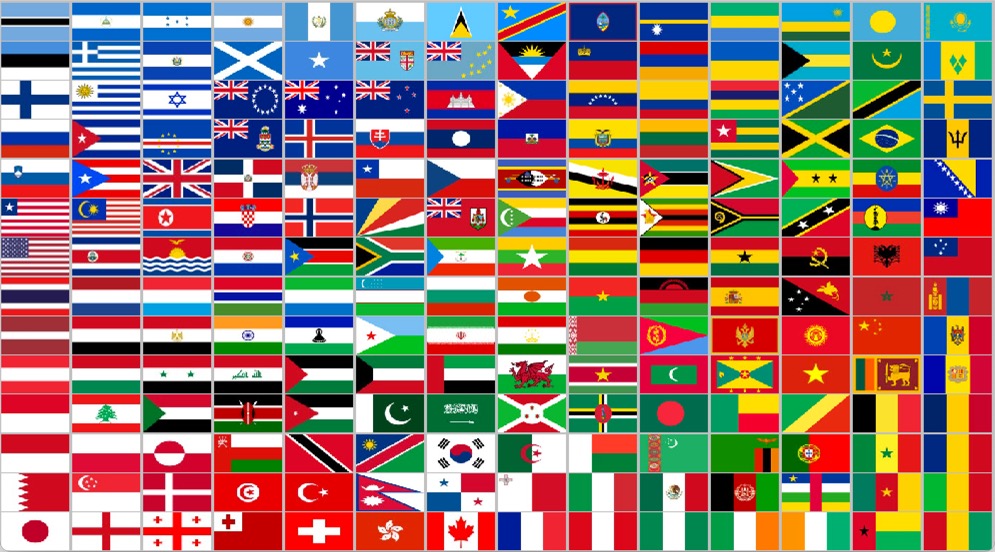}
        \end{subfigure}
      \hfill
      \begin{subfigure}{0.49\linewidth}
        \centering
        \includegraphics[width=\linewidth]{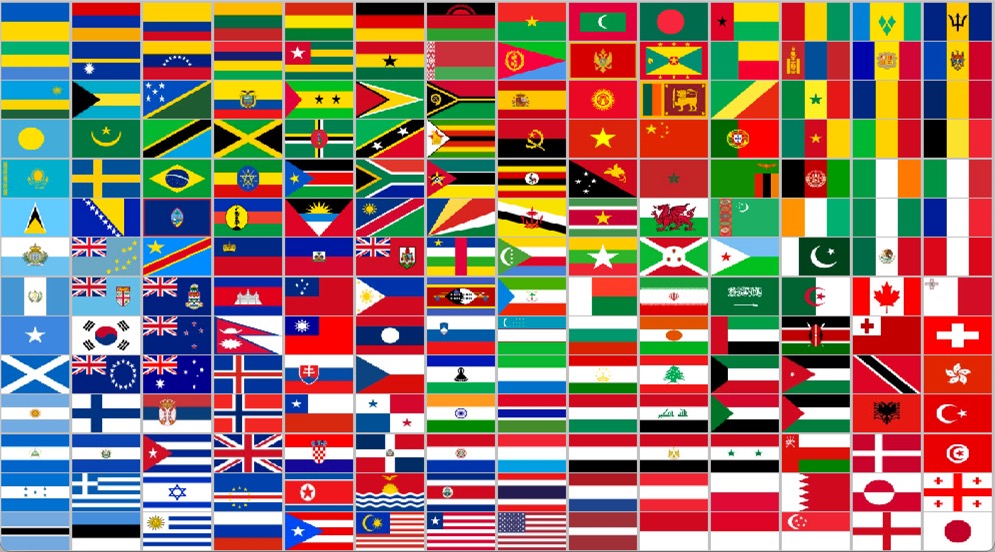}
      \end{subfigure}
    \caption{
    Flags arranged by their similarity. On the left, the American flag was pinned to the middle of the left side. on the right, it was pinned at the center of the bottom line.}
    \label{fig:flags}
\end{figure}

Sometimes it is desirable to keep some images fixed at certain positions (see Figure \ref{fig:flags}). 
This is possible with two minor changes to the algorithm: In a first step, the images or the corresponding vectors are assigned to the desired positions. These positions are then never changed again. Also, an additional weighting factor is introduced for filtering, where the fixed positions are weighted more. This results in neighboring map vectors becoming similar to these fixed vectors, which in turn results in similar images also being placed nearby. The rest of the algorithm remains the same. 

\section{Conclusions}
We presented a new evaluation metric to assess the quality of grid-based image arrangements. 
The basic idea is not to evaluate the preservation of the HD neighbor ranks of an arrangement, but the preservation of the average distances of the neighborhood. 
Furthermore, we do not weight all distances equally either, because for humans the preservation of small distances seems to be more important than that of larger distances. 
User experiments have shown that distance preservation quality better represents human-perceived qualities of sorted arrangements than other existing metrics. 
If the overall impression of an arrangement is to be evaluated, the $DPQ_p^-$ metric had a small advantage. For predicting how fast images can be found for an arrangement, $DPQ_p$ was better. In general, however, these differences are small and we recommend always using $DPQ_p$. 

Large $p$ values lead to the highest correlations with user perception.
This implies that for a "good" arrangement it is essentially important that the sum of the distances to the immediate neighbors on the 2D grid is as small as possible. 
The same is true for one-dimensional sorting, which is "optimal" only if the sum of the differences to the direct neighbors is minimal. 
It remains to be investigated whether distance preservation quality is a useful metric for evaluating the quality of other, non-grid-based dimensionality reduction methods. 

Furthermore, we have presented \emph{Linear Assignment Sorting} which is a simple but at the same time very effective sorting method. It achieves very good arrangements according to the new metric as well as for other metrics. The \emph{Fast Linear Assignment Sorting} variant can achieve better arrangements than existing methods with reduced complexity. 

The ideas presented in this paper can be developed in numerous directions.
Currently, the filtering approach optimizes for mean HD distances for equal 2D distances. We will investigate whether further improvement is possible by exploiting the fact that for equal 2D distances on the grid, the sorted HD distances better represent the perceived quality. 
Currently, our sorting method only supports regular grids. We want to investigate how this approach can be extended to densely packed rectangles of various sizes.

%
%
%

\bibliographystyle{unsrt}
\bibliography{references}

\end{document}